\documentclass{article}

\usepackage[margin=1in]{geometry}

\usepackage{microtype}
\usepackage{graphicx}
\makeatletter
\def\maxwidth#1{\ifdim\Gin@nat@width>#1 #1\else\Gin@nat@width\fi}
\makeatother
\usepackage{subcaption}
\usepackage{booktabs}
\usepackage{amsmath}
\usepackage{amssymb}
\usepackage{amsfonts}
\usepackage{multirow}
\usepackage{verbatim}
\usepackage{caption}
\usepackage{longtable}
\usepackage{float}
\usepackage{CJKutf8} %
\usepackage{kotex}
\usepackage{threeparttable}
\usepackage{vwcol}

\usepackage{mathptmx}
\newcommand{\armenian}{\fontencoding{OT6}\fontfamily{cmr}\selectfont}
\DeclareTextFontCommand{\textarmenian}{\armenian}
\usepackage[OT6,T1]{fontenc}

\usepackage{enumitem}
\usepackage{tablefootnote}
\usepackage[authoryear, sort&compress, round]{natbib}
\usepackage[dvipsnames]{xcolor}
\usepackage{xspace}
\usepackage{textcomp}
\usepackage{makecell}
\usepackage{lscape} 
\usepackage{siunitx}
\usepackage{geometry}
\usepackage[most]{tcolorbox}
\usepackage[bottom]{footmisc}

\usepackage{amssymb}%
\usepackage{pifont}%
\usepackage{scrextend}

\newcommand{\IGNORE}[1]{}

\usepackage{hyperref}

\definecolor{darkblue}{rgb}{0, 0.2, 0.7}
\hypersetup{
    colorlinks = true,
    linkcolor = darkblue,
    anchorcolor = darkblue,
    citecolor = darkblue,
    filecolor = darkblue,
    urlcolor = darkblue
}

\usepackage{soul}
\usepackage[dvipsnames]{xcolor}

\usepackage[colorinlistoftodos,prependcaption,textsize=tiny]{todonotes}

\usepackage{xspace}

\makeatletter
\DeclareRobustCommand\onedot{\futurelet\@let@token\@onedot}
\def\@onedot{\ifx\@let@token.\else.\null\fi\xspace}

\makeatother

\newcounter{sm}

\newcounter{adai}

\newcounter{ruder}

\newcounter{kevinrobinson}

\usepackage{appendix}
\usepackage{cleveref}
\crefname{section}{Section}{Sections}
\crefname{subsection}{Section}{Sections}
\crefname{table}{Table}{Tables}
\crefname{figure}{Figure}{Figures}
\crefname{algorithm}{Algorithm}{}
\crefname{equation}{eq.}{}
\crefname{appendix}{Appendix}{}
\usepackage{multicol}

\definecolor{lightgray}{rgb}{0.95, 0.95, 0.95}
\definecolor{darkgray}{rgb}{0.4, 0.4, 0.4}
\definecolor{editorGray}{rgb}{0.95, 0.95, 0.95}
\definecolor{editorOcher}{rgb}{1, 0.5, 0} %
\definecolor{editorGreen}{rgb}{0, 0.5, 0} %
\definecolor{orange}{rgb}{1,0.45,0.13}		
\definecolor{olive}{rgb}{0.17,0.59,0.20}
\definecolor{brown}{rgb}{0.69,0.31,0.31}
\definecolor{purple}{rgb}{0.38,0.18,0.81}
\definecolor{lightblue}{rgb}{0.1,0.57,0.7}
\definecolor{lightred}{rgb}{1,0.4,0.5}
\usepackage{upquote}
\usepackage{listings}
\lstdefinelanguage{CSS}{
  keywords={color,background-image:,margin,padding,font,weight,display,position,top,left,right,bottom,list,style,border,size,white,space,min,width, transition:, transform:, transition-property, transition-duration, transition-timing-function},	
  sensitive=true,
  morecomment=[l]{//},
  morecomment=[s]{/*}{*/},
  morestring=[b]',
  morestring=[b]",
  alsoletter={:},
  alsodigit={-}
}

\lstdefinelanguage{JavaScript}{
  morekeywords={typeof, new, true, false, catch, function, return, null, catch, switch, var, if, in, while, do, else, case, break},
  morecomment=[s]{/*}{*/},
  morecomment=[l]//,
  morestring=[b]",
  morestring=[b]'
}

\lstdefinelanguage{HTML5}{
  language=html,
  sensitive=true,	
  alsoletter={<>=-},	
  morecomment=[s]{<!-}{-->},
  tag=[s],
  otherkeywords={
  >,
	<!DOCTYPE,
  </html, <html, <head, <title, </title, <style, </style, <link, </head, <meta, />,
	</body, <body,
	</div, <div, </div>, 
	</p, <p, </p>,
	</script, <script,
  <canvas, /canvas>, <svg, <rect, <animateTransform, </rect>, </svg>, <video, <source, <iframe, </iframe>, </video>, <image, </image>, <header, </header, <article, </article
  },
  ndkeywords={
  =,
  charset=, src=, id=, width=, height=, style=, type=, rel=, href=,
  fill=, attributeName=, begin=, dur=, from=, to=, poster=, controls=, x=, y=, repeatCount=, xlink:href=,
  margin:, padding:, background-image:, border:, top:, left:, position:, width:, height:, margin-top:, margin-bottom:, font-size:, line-height:,
  transform:, -moz-transform:, -webkit-transform:,
  animation:, -webkit-animation:,
  transition:,  transition-duration:, transition-property:, transition-timing-function:,
  }
}

\lstdefinestyle{htmlcssjs} {%
  basicstyle={\footnotesize\ttfamily},   
  frame=b,
  xleftmargin={0.75cm},
  numbers=left,
  stepnumber=1,
  firstnumber=1,
  numberfirstline=true,	
  identifierstyle=\color{black},
  keywordstyle=\color{blue}\bfseries,
  ndkeywordstyle=\color{editorGreen}\bfseries,
  stringstyle=\color{editorOcher}\ttfamily,
  commentstyle=\color{brown}\ttfamily,
  language=HTML5,
  alsolanguage=JavaScript,
  alsodigit={.:;},	
  tabsize=2,
  showtabs=false,
  showspaces=false,
  showstringspaces=false,
  extendedchars=true,
  breaklines=true,
  literate=%
  {Ö}{{\"O}}1
  {Ä}{{\"A}}1
  {Ü}{{\"U}}1
  {ß}{{\ss}}1
  {ü}{{\"u}}1
  {ä}{{\"a}}1
  {ö}{{\"o}}1
}
\lstdefinestyle{py} {%
language=python,
literate=%
*{0}{{{\color{lightred}0}}}1
{1}{{{\color{lightred}1}}}1
{2}{{{\color{lightred}2}}}1
{3}{{{\color{lightred}3}}}1
{4}{{{\color{lightred}4}}}1
{5}{{{\color{lightred}5}}}1
{6}{{{\color{lightred}6}}}1
{7}{{{\color{lightred}7}}}1
{8}{{{\color{lightred}8}}}1
{9}{{{\color{lightred}9}}}1,
basicstyle=\footnotesize\ttfamily, %
numbers=left,               %
numbersep=5pt,              %
tabsize=4,                  %
extendedchars=true,         %
breaklines=true,            %
keywordstyle=\color{blue}\bfseries,
frame=b,
commentstyle=\color{brown}\itshape,
stringstyle=\color{editorOcher}\ttfamily, %
showspaces=false,           %
showtabs=false,             %
xleftmargin=17pt,
framexleftmargin=17pt,
framexrightmargin=5pt,
framexbottommargin=4pt,
showstringspaces=false,      %
}%

\usepackage{ulem}               %
\colorlet{strikeColor}{BrickRed}
\colorlet{suggestColor}{Black}

\DeclareMathOperator*{\argmin}{arg\,min}

\usepackage{devanagari}

\title{\vspace{-2em}%
  \hrule height 4pt%
  \vskip 0.25in%
  \vskip -\parskip%
  \textbf{Scaling Laws of Motion Forecasting and Planning \newline Technical Report}
  \vskip 0.2in%
  \vskip -\parskip%
  \hrule height 1pt%
  \vskip 0.09in}

\author{Mustafa Baniodeh\thanks{Equal contributions.}, \ 
Kratarth Goel\footnotemark[1], \ 
Scott Ettinger, \ 
Carlos Fuertes, \ 
Ari Seff, \ 
Tim Shen, \\
Cole Gulino, \
Chenjie Yang, \
Ghassen Jerfel, \ 
Dokook Choe, \ 
Rui Wang, \
Benjamin Charrow, \\
Vinutha Kallem, \ 
Sergio Casas,
Rami Al-Rfou, \ 
Benjamin Sapp, \ 
Dragomir Anguelov\\
~ \\
Waymo LLC\footnote{Please send correspondence to \texttt{\{kratarth,bensapp\}@waymo.com}. See Section~\ref{sec:acknowledgements} 
for further acknowledgements.}}

\date{}

\begin{document}
\maketitle

\begin{abstract}
We study the empirical scaling laws of a family of encoder-decoder autoregressive transformer models on the task of joint motion forecasting and planning in the autonomous driving domain. Using a $\sim500$ thousand hours driving dataset, we demonstrate that, similar to language modeling, model performance improves as a power-law function of the total compute budget, and we observe a strong correlation between model training loss and model evaluation metrics. Most interestingly, closed-loop metrics also improve with scaling, which has important implications for the suitability of open-loop metrics for model development and hill climbing. We also study the optimal scaling of the number of transformer parameters and the training data size for a training compute-optimal model. We find that as the training compute budget grows, optimal scaling requires increasing the model size 1.5x as fast as the dataset size. We also study inference-time compute scaling, where we observe that sampling and clustering the output of smaller models makes them competitive with larger models, up to a crossover point beyond which a larger models becomes more inference-compute efficient. Overall, our experimental results demonstrate that optimizing the training and inference-time scaling properties of motion forecasting and planning models is a key lever for improving their performance to address a wide variety of driving scenarios. Finally, we briefly study the utility of training on general logged driving data of other agents to improve the performance of the ego-agent, an important research area to address the scarcity of robotics data for large capacity models training.

\end{abstract}

\newpage
\newgeometry{top=20mm, bottom=20mm} 
\tableofcontents %
\restoregeometry
\newpage

\section{Introduction}
\label{sec:introduction}
Motion forecasting and planning are core autonomous vehicle capabilities. The forecasting task typically involves predicting the likely future
trajectories of dynamic agents (e.g., pedestrians, cyclists, and vehicles). The planning task predicts trajectories for the autonomous vehicle (the ego-agent) conditioned on the route objective, optimizing for comfortable and safe motion with respect to the likely behaviors of other dynamic agents. The complexity of this
task arises from the inherent uncertainty in predicting the dynamic agents'
behavior, the complex interactions between them, and the need to reason about
the long-term consequences of actions in a continuous state space.
Systematically improving the performance on this task requires advances in modeling
to account for the interactions between dynamic agents, thoughtful 
model input design to reduce information bottlenecks due to limitations in the interface
with the perception system, and training data scaling to improve the model performance on long-tail scenarios.

The fact that  past and future agent trajectories are temporal sequences makes
transformers well-suited for the motion forecasting task ~\citep{girgis2022latentvariablesequentialset, 9197340, ngiam2022scenetransformerunifiedarchitecture, yu2020spatiotemporalgraphtransformernetworks,Yuan_2021_ICCV, 2022Wayformer}. \citet{2022Wayformer} showed that an encoder-decoder transformer
setup simplifies the design space and enables its systematic study,
especially with regards to the different options for fusing the input modalities. \citet{2022Wayformer} was applied to the marginal a-posteriori trajectory prediction task, i.e., the prediction of multiple possible futures for each agent independent from the futures of the remaining agents. Several works studied autoregressive transformers to jointly predict the future trajectories of
multiple or all agents in the scene~\citep{ngiam2022scenetransformerunifiedarchitecture, Seff2023MotionLM,jia2024ampautoregressivemotionprediction,  shi2024mtrmultiagentmotionprediction, zhou2024behaviorgptsmartagentsimulation}. Jointly predicting multiple agents is a
simple form of world modeling, which can be used to study how increasing the scene modeling complexity can benefit the ego-agent planning task. In
this report, we study the improvements in the joint prediction task as we
systematically increase the compute, data, and model size, using the
architecture from \citet{Seff2023MotionLM}.

In recent years, it has been repeatedly shown that the performance of deep
learning models scales predictably with data and compute, as first highlighted
and studied in ~\citep{hestness2017deep}. \citet{kaplan2020scaling}
systematically studied the improvements in the cross-entropy loss of next-token
prediction language models as the model size, dataset size, and training
compute were scaled, and found a power-law relationship that empirically holds
for 7 orders of magnitude. This provided strong evidence that warranted calling
it an empirical "law". The work also proposed a methodology to find the optimal allocation
of a compute budget for increasing the model size versus the training dataset
size. \citet{henighan2020scaling} found the same relationship to hold for
transformer decoders on the autoregressive generation task in several other
modalities, further supporting the idea that these power-law relationships
could be universal laws independent of the data modality of the generative task.

Most principled studies of scaling laws exist only for language models trained
on internet-scale text datasets, with only a few notable exceptions. Without
firm theoretical explanations for this empirical phenomenon, it is not
immediately clear that these laws should hold for the joint-prediction motion
forecasting task, and how such improvement correlate with close-loop and real world models performance. To this end, in this report, we carefully construct a study
with the aim of answering the following key questions:

\begin{enumerate}
    \item \textbf{Pre-training Scaling Laws}
        \begin{enumerate}
            \item How does the cross-entropy loss scale as we increase the model size, dataset size, and compute in tandem?
            \item What is the optimal model and dataset scaling as we scale training compute?
            \item Do downstream metrics relevant to driving follow improvements in the cross-entropy loss?
        \end{enumerate}
    \item \textbf{Closed-loop Scaling Laws}: Do closed-loop planning metrics (within a realistic simulated environment) correlate with pre-training cross-entropy? In other words, are larger models safer and more competent drivers?
    \item \textbf{Inference Scaling Laws}: How does performance change as we increase inference-time model sampling? Can smaller models be competitive with larger models? 
    \item \textbf{Cross-Agent Skills Transfer}: Can training on "passive" driving logs help train larger models that translate to better AV ego-agents?
\end{enumerate}

Following \citet{hoffmann2022training}, we conduct an iso-FLOP analysis to find the compute-optimal models across various fixed training compute budgets. We find that the cross-entropy loss for our joint prediction formulation of motion forecasting also follows a power-law relationship as a function of the training compute. Furthermore, the optimal allocation of compute should grow 1.5x as fast as the dataset size. Interestingly, at the same training compute budget, an optimal LLM is $\sim50$ times larger than an optimal motion forecasting model. We hypothesize that this, at least in part, can be attributed to driving data distributions; i.e. models need more training data to capture less common driving modes. To our knowledge, this is the first comprehensive study showing that the optimal number of model parameters could be drastically different for driving tasks compared to language models -- potentially suggesting the importance of collecting more data, or perhaps improving the training data sampling techniques for this domain. This finding also indicates that improvement in model performance by scaling training compute can be directly leveraged to improve the performance of onboard systems, as the optimal models are relatively small in size.

We validate that the improvement in the model training loss leads to consistent improvements in the open-loop precision and coverage metrics. For autonomous driving, while open-loop metrics are important indicators of model performance, it is unclear if they are an unbiased estimator of closed-loop simulation performance for driving agents navigating in real scenarios. Conventional wisdom in the field, as well as previous studies such as~\citep{dauner2023nuplan}, suggest that open- and closed-loop evaluations are misaligned. To study the correlation between cross-entropy loss, open-loop metrics, and closed-loop simulation, we also conduct a scaling analysis showing improvements in closed-loop metrics that are consistent with the loss improvements. These evaluation results suggest that performance on the driving task can be improved by pursuing compute and data scaling for the supervised training task.

We study inference scaling to answer questions about model performance as a function of the inference FLOPs. We observe that increasing the inference compute by scaling the sampling of smaller models makes them competitive with larger models, up to a crossover point where even limited sampling from a larger model is more inference compute-efficient.

One of the major challenges to scaling robotic manipulation deep learning models is the difficulty and cost of collecting human demonstration data. There are many ongoing investigations to leverage generic internet-scale video datasets to scale pre-training~\citep{ye2024latentactionpretrainingvideos, cheang2024gr2generativevideolanguageactionmodel, wu2023unleashinglargescalevideogenerative}. For driving, it might be relatively easier to passively collect driving logs, which can be used to scale training. At the end of this report, we present a preliminary study of skills transfer from observed driving logs to the AV ego-agent.

In Section~\ref{sec:model}, we briefly present the motion forecasting problem formulation and model architecture. For further details, we refer the reader to more detailed works on this topic~\citep{2022Wayformer, Seff2023MotionLM, Ettinger_2021_ICCV}. Section~\ref{sec:dataset} introduces the training dataset. In Section~\ref{sec:scaling_laws}, we present our scaling laws analysis, followed by open-loop and closed-loop evaluations in Section~\ref{sec:evaluation}. The inference scaling study is presented in Section~\ref{sec:inference}, and Section~\ref{sec:transfer} presents the cross-agent skills transfer study. We conclude with a review of related work in Section~\ref{sec:related_work} and a discussion of the limitations and implications of the results in Section~\ref{sec:discussion}.

\section{Motion Forecasting Modeling}
\label{sec:model}
\subsection{Problem Formulation}
 
We formulate the problem of motion forecasting as a conditional sequence generation problem similar to modern language models. 
Given a set of perception features representing the scene context over the past few seconds, the task is to generate the future motion tokens for $M$ agents in the scene, which we refer to as \textit{agents of interest} or \textit{modeled agents}. An agent of interest can be the autonomous vehicle, another vehicle, a pedestrian, or a cyclist.

The scene context $S$ consists of multi-modal data including road information, traffic
light state history, and agent state history. The histories are provided for $T_\text{history}$ time steps (the current time step and $T_\text{history}-1$ past time steps).

\textbf{Agent History} contains sequences of past states for $S_a$ contextual agents. This input has shape $[S_{a}, T_\text{history}, D_{a}]$. For each time step, we consider
features that define the state of the agent: $xyz$ position, heading, velocity, and bounding box extents. Note that the $M$ modeled agents are a subset of the $S_a$ contextual agents.

\textbf{Roadgraph} represents the road shape around the autonomous vehicle with a collection of $S_{r}$ line segments
specified by their endpoints and annotated with type information. This input has shape $[S_{r}, 1, D_{r}]$. Each polyline has endpoint $xyz$ position, direction, type and validity features. Note that there is no time
dimension for the road features, but we include a time dimension of 1 for
homogeneity with the other modalities.

\textbf{Traffic Light State History} describes the position, state, and confidence of the state estimation over time for $S_{tls}$ traffic lights, in a feature tensor of shape $[S_{tls}, T_\text{history}, D_{tls}]$.

The motion forecasting task is to generate joint agent actions $A_{t}= \{a^1_{t}, a^2_{t}, ..., a^M_{t} \}$
for $M$ agents of interest at future timesteps $t = 1, ..., T$. 
Each agent action $a^m_{t}$ is a discrete motion token representing a two-dimensional Verlet-wrapped displacement in Bird's-Eye-View, as described in \citep{Seff2023MotionLM}. From these actions, agent trajectories $Y = \{Y_1, \dots, Y_T \}$ can be calculated easily via integration of the displacements over the discrete time steps.

\subsection{Modeling}

Large Language Models (LLMs) are pre-trained to maximize the probability of the next token conditioned on the previous text \citep{brown2020languagemodelsfewshotlearners, hoffmann2022training}. This approach has
found success in continuous domains such as speech and image
generation, as well. Leveraging the flexibility of arbitrary categorical distributions, we can represent continuous data with a set of discrete tokens, reminiscent of
language model vocabularies.

In driving scenarios, road users may be likened to participants in a constant
dialogue, continuously exchanging a dynamic series of actions and reactions
mirroring the fluidity of communication. Navigating this rich web of
interactions requires the ability to anticipate the likely maneuvers and
responses of the involved actors. Just as today’s language models can capture
sophisticated distributions over conversations, we leverage similar sequence
models to forecast the behavior of road agents. We follow the
~\citep{Seff2023MotionLM} design to model the task of trajectory prediction as
an auto-regressive sequence prediction problem over discrete action spaces much
like the objective of large language models.

\subsubsection{Joint agent modeling}
The task of the motion prediction model is to produce joint future actions $A_{t}$ of the modeled agents, as explained above. Given the future is inherently multi-modal depending on complex interactions between scene elements, extending the problem definition to predict full future joint
rollouts of all scene elements (world modeling) allows the model to reason about richer representations. It is also a general, foundational representation which can be readily adapted to a variety of important related tasks including marginal, joint or conditional predictions for behavior prediction, sim agents or planning applications.

In our modeling framework, we sample a predicted action for each modeled agent
at each future time step. These actions are formulated as discrete motion tokens
from a finite vocabulary. 
We factorize
the distribution over joint future action sequences as a product of
conditionals:

\begin{equation} p_{\theta}(A_1, A_2, ..., A_T \mid  S) = \prod_{t=1}^T
p_{\theta} (A_t \mid A_{\textless t},  S) \\ \end{equation}

\begin{equation}\label{eqn:joint_agent} p_{\theta}(A_t \mid A_{\textless t},  S)
= \prod_{m=1}^M p_{\theta} (a^m_t \mid a_{\textless t}^1, \dots, a_{\textless t}^M , S) \end{equation}

Equation \ref{eqn:joint_agent} represents the fact that we treat agent actions
as conditionally independent at time $t$, given the previous actions and scene
context. 
During training, we use teacher forcing and minimize the cross entropy loss between the predicted and target discrete motion tokens, thus maximizing the likelihood of multi-agent actions expressed above. We train all models in this study with $M=8$ agents.

\subsection{Model Architecture}

We follow the setup from MotionLM \citep{Seff2023MotionLM}. We prioritize design choices that maximize the scalability of our model
development process.
Our model consists of two main networks: an encoder which processes initial
scene elements, and a motion decoder which performs both cross-attention to
the scene encodings and self-attention along agent motion tokens.

\begin{figure}[ht]
    \centering
    \includegraphics[trim={0 0cm 0cm 0cm},width=1.0\linewidth]{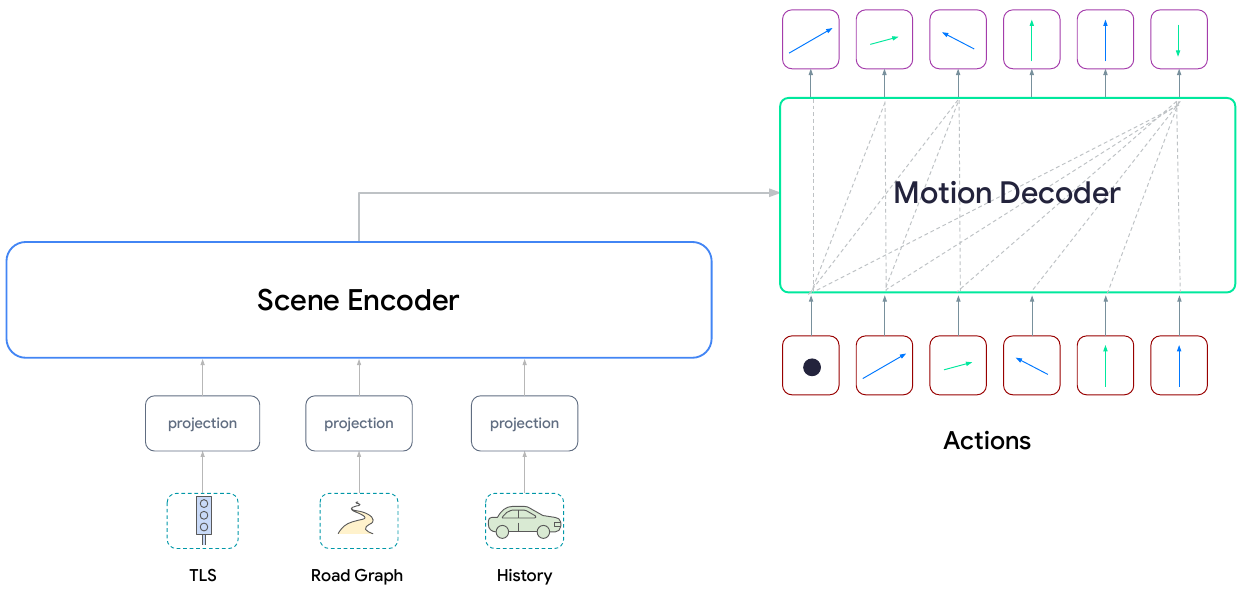}
    \caption{The model architecture is a pair of encoder/decoder Transformer networks. This model takes multimodal scene data as input and produces a series of motion tokens autoregressively.}
    \label{fig:wayformer}
\end{figure}

\subsubsection{Scene Encoder}
The scene encoder processes information from multiple input modalities,
including the roadgraph, traffic light states, and agents trajectory histories.
Here, we follow the design of the early fusion network proposed
by~\citep{2022Wayformer} as the backbone of our scene encoder. Early fusion is
favored for its flexibility to process all modalities together with minimal
inductive bias. For additional details, refer to~\citep{2022Wayformer}. The scene encoder encodes all the modalities in a global frame of reference defined by the autonomous vehicle pose. This stands in contrast to Wayformer \citep{2022Wayformer} and MotionLM \citep{Seff2023MotionLM}, where the scene encoding was done for each modeled agent in its own local frame of reference.

\subsubsection{Motion Decoder}
Our motion decoder is tasked with generating sequences of motion tokens for
multiple agents by cross-attending to the outputs of the scene encoder.

\textbf{Discrete motion tokens:} We elect to transform trajectories comprised
of continuous waypoints into sequences of discrete tokens $a^m_t$. This enables
treating sampling purely as a classification task at each timestep.
Discretizing continuous targets in this manner has proven effective in other
inherently continuous domains, e.g., in audio generation \citep{yang2024uniaudioaudiofoundationmodel}, mesh generation \citep{hao2024meshtronhighfidelityartistlike3d} and
robotic manipulation \citep{kim2024openvlaopensourcevisionlanguageactionmodel}. We suspect that discrete motion tokens also
naturally hide some precision from the model, possibly mitigating compounding
error effects that could arise from imperfect continuous value prediction \citep{pertsch2025fastefficientactiontokenization}.

\textbf{Flattened agent-time self-attention:} While separate passes of
factorized agent and time attention are also possible~\citep{ngiam2022scenetransformerunifiedarchitecture, 2022Wayformer, jia2024ampautoregressivemotionprediction}, we use a single pass of joint agent-time self-attention
here for simplicity.
So, given a target sequence of length $T$ for each of
$M$ agents, we perform self-attention over $MT$ elements, with appropriate
causal masking to avoid future leakage.

This motion decoder follows that of MotionLM \citep{Seff2023MotionLM} closely, with the exception that here all motion tokens cross-attend to the shared scene encoder tokens, and in MotionLM tokens from different agents cross-attended separate scene encoder tokens as the scene encoding was done per agent.

\section{Dataset}
\label{sec:dataset}
Data is a key enabler for scaling. Models scale with compute and capacity only when unconstrained by amount of data \citep{kaplan2020scaling}. More so, recently publicly available research finds that more data for a fixed amount of training compute, or overtraining of models of a fixed capacity results in further improvements \citep{devries2023chinchilla_analysis}. In addition, improvements to both the quantity and quality of the data results in significant improvements across foundation model generations. For example, the number of tokens used increased from 1.8T to 14T tokens from Llama2 to Llama3 family of models \citep{grattafiori2024llama3herdmodels} for training the flagship 405B parameter model. These improvements include the development
of more careful pre-processing and curation pipelines for pre-training data and the development of more
rigorous quality assurance and filtering approaches for post-training data. 

We curate a dataset following the trends above.  As a quality measure, we utilize a dataset consisting of safety driver demonstrations; no semi- or fully- autonomous miles were used in this study.  We sample nearly 6 million unique runs, and sample 30 second run segments within those. As part of this sampling, we perform simple filtering and deduplication to ensure the data is valid, interesting and diverse. These scenarios are composed of a rich mixture of real world driving situations like driving through busy intersections in dense urban cities (e.g., San Francisco, Phoenix, Los Angeles), dealing with construction zones and road closures, interacting and safely navigating around emergency vehicles, driving on higher speed limit areas like freeways, complex you-go-I-go interactions with pedestrians and cyclists. We summarize some interesting statistics of the dataset in table \ref{table:data_table}.

\begin{table}[h!]
\centering
\begin{tabular}{|c l|} 
 \hline
 \textbf{Data metric} & \textbf{Value} \\ [0.5ex] 
 \hline
 Number of run segments & 59.8 million \\ 
 Number of agents & 373 billion \\
 Number of hours of driving & 447 thousand \\
 Number of miles of driving & 5.6 million \\
 Number of training examples & 541 million \\
 \hline
\end{tabular}
\caption{Statistics about the data used for the scaling analysis.}
\label{table:data_table}
\end{table}

As explained in \Cref{sec:model} our training examples consist of driving history and future prediction. For this study we use 5 seconds history to predict for 11 seconds of future. The long horizon prediction, while challenging for the task of motion prediction, allows for modeling complex inter-agent dynamic and static interactions. We use overlapping sliding windows to construct multiple training examples from the 30 second run segments. We currently use a sliding window of 1.5 seconds to create multiple examples from each run segment.

\section{Scaling Laws}
\label{sec:scaling_laws}
We investigate the scaling behavior of our transformer model on the multi-agent joint future trajectory prediction task. We systematically vary the model size, dataset size, and
training compute to address two key questions: 1) Does the cross-entropy loss follow
power-law scaling as observed in language and other domains? 2) With a fixed
compute budget, what is the optimal allocation between increasing model size
and dataset size?

Following~\citet{kaplan2020scaling, hoffmann2022training}, we parameterize the
training loss as $L(N,D)$, where $N$ represents the total number of transformer
parameters (excluding embedding layers, see \ref{appendix:flops}) and $D$ is
the number of training examples.  Our model employs an encoder to transform the model inputs into a scene embedding, 
as well as a decoder for auto-regressive prediction of motion action tokens for the dynamic agents. Since the
scene embedding dimensions and motion tokens are fixed per training example, we
parameterize the loss using the number of training examples, deviating from the
common practice in decoder-only models which use the total number of training
tokens.

To determine the optimal model and data scaling, we
follow~\citet{hoffmann2022training, kaplan2020scaling} and aim to minimize the
loss function subject to a fixed compute budget constraint, FLOPs$(N, D) = C$,
\begin{equation}\label{eq:optimal} N_{opt}, D_{opt} = \underset{N, D \text{
s.t. } FLOPs(N, D) = C}{\text{argmin}} L(N, D) \end{equation}

We observe that, for our symmetric encoder-decoder models, $N_{opt} \propto
C^{0.63}$ and $D_{opt} \propto C^{0.44}$. This indicates that for optimal
training compute efficiency, the optimal model size should grow $\sim1.5$ times as fast as
the number of training examples. We report the analysis details next.
\subsection{Estimating Optimal Model and Data Scaling}

\begin{figure}[h!]
    \centering
    \includegraphics[scale=0.45]{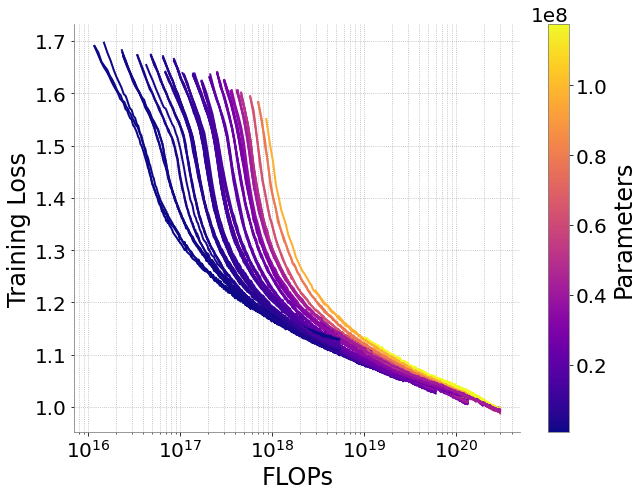}
    \caption{Training cross-entropy loss for all experiments as a function of
    compute. The losses are presented without smoothing.}\label{fig:loss}
\end{figure}

The cross-entropy loss for encoder-decoder transformer models can be parameterized in terms
of their number of encoder and decoder parameters~\citep{ghorbani2021scaling},
\begin{equation*}
  L(N_e, N_d, D) = E  + \left(\frac{F}{N_e}\right)^{\alpha_e}\left(\frac{G}{N_d}\right)^{\alpha_d} + \frac{B}{D^\beta}
\end{equation*}
where $N_e$ and $N_d$ are the number of encoder and decoder parameters,
respectively. $\alpha_e$ and $\alpha_d$ are encoder and decoder-specific
exponents. $F$, $G$ and $B$ are normalization constants. $E$ is interpreted as the entropy of the data distribution~\citep{henighan2020scaling}. As mentioned earlier, $D$ is the number of training examples.

We train models that are symmetric in their number of encoder and decoder
layers. Consequently, our decoders hold $\frac{4}{7}$ of the total number of
non-embedding parameters (see Appendix \ref{appendix:flops} for parameter
estimates). Thus, our encoder-decoder models can be parameterized similarly
to \citep{hoffmann2022training}:
\begin{equation}
  L(N, D) = E + \frac{A}{N^\alpha} + \frac{B}{D^\beta}
\end{equation}
This allows us to study model size in terms of the total number of model
parameters, analogous to decoder-only models. As explained in the previous
section, $D$ in our case represents the total number of training examples. $A$ and $B$ are normalization constants which can also be obtained empirically by fitting direct fit to the data (see approach 3 in~\citet{hoffmann2022training}).

\begin{figure}[h!]
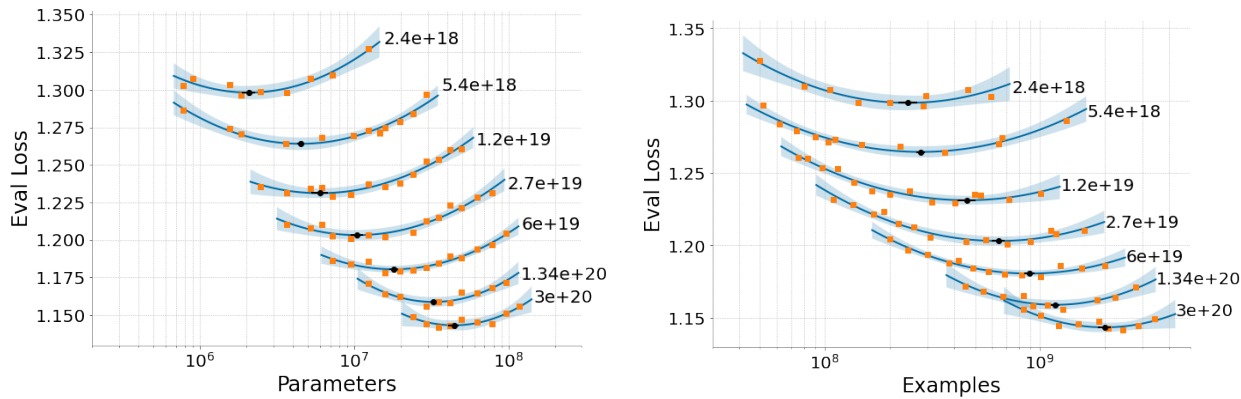

    \centering
    \begin{subfigure}{0.49\textwidth}
        \includegraphics[width=\textwidth]{figures/bands_parameters.pdf}
    \end{subfigure}
    \hfill
    \begin{subfigure}{0.49\textwidth}
        \includegraphics[width=\textwidth]{figures/bands_data.pdf}
    \end{subfigure}
    \caption{Final losses as a function of the total number of
    non-embedding parameters (left), and the total number of training examples (right), grouped into iso-FLOP bands. The lines represent parabolic fits, with error bands indicating 3 standard deviations propagated from the fit parameters covariance (see Appendix~\ref{appendix:fits_and_errors}). The minimas of the parabolas indicate the optimal model and dataset sizes at a fixed compute budget.}\label{fig:loss_bands}
\end{figure}

We train 84 models ranging in size from 900K to 118M parameters across seven
compute budgets spanning more than two orders of magnitude. To obtain models with different sizes, we vary parameter counts by equally increasing the number of encoder and decoder layers, while keeping width-to-depth ratio at either 8 or 16. To get iso-FLOP bands, we vary training compute via number of model parameters and training steps. The batch-size was fixed to 512 examples. With 673 context tokens and 176 predicted action tokens, the batch has 434K tokens. We employ AdamW
optimization with a fixed batch size throughout our study. A cosine learning
rate schedule is used, with 3000 warmup steps, a peak value of $2 \times
10^{-4}$, and a decay to $2 \times 10^{-5}$ fixed to the total number of
training steps. Training is performed using a cross-entropy loss as previously
described. Figure~\ref{fig:loss} presents the training loss curves for all
experiments as a function of the training budget.

\begin{figure}[h!]
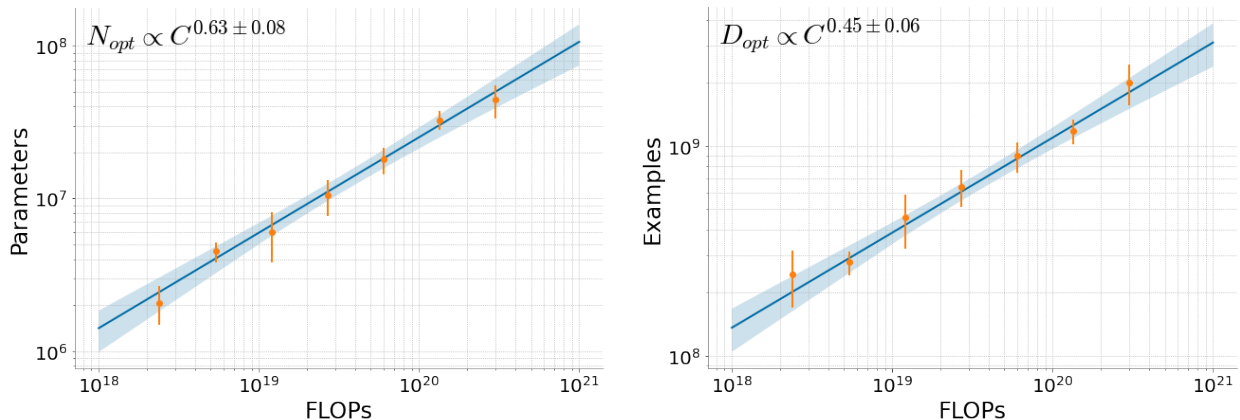

    \centering
    \begin{subfigure}{0.49\textwidth}
        \includegraphics[width=\textwidth]{figures/optimal_parameters.pdf}
    \end{subfigure}
    \hfill
    \begin{subfigure}{0.49\textwidth}
        \includegraphics[width=\textwidth]{figures/optimal_data.pdf}
    \end{subfigure}
    \caption{Training-compute optimal number of model parameters ($N_{opt}(C)$)
    and number of training examples ($D_{opt}(C)$) as a function of total compute.
  Data points are obtained from the parabola fits, with error bars representing
3 standard deviation. The lines correspond to power-law
fits of the form $ax^b$, and the resulting exponent values are displayed. The
error bands indicate 3 standard deviations propagated from the fit parameters
covariance (see \Cref{appendix:fits_and_errors}).}
    \label{fig:optimal_params_data}
\end{figure}

In order to find out the optimal number of model parameters and training examples, we plot the final losses on a validation dataset grouped in iso-FLOP bands in Figure~\ref{fig:loss_bands}. We fit the bands using a parabola parameterized as:
\begin{equation*}
L_C(x) = a\left(\log x - \log x_C^{opt}\right)^2 + L_C^{opt}
\end{equation*}
where $x$ is either number of parameters or number of data points. $a$, $x_C^{opt}$, and $L_C^{opt}$ are the fit
parameters. The latter two are the optimal values we are extracting for each compute budget $C$. We report the 3 standard deviations in $N_{opt}$, $D_{opt}$, and $L_{opt}$
directly from the fit parameter variances.  Figure~\ref{fig:optimal_params_data} shows the $N_{opt}(C)$ and $D_{opt}(C)$ obtained
from the parabola fits, along with a power-law fit of the form $ax^{b}$.
Finally, we derive the exponents $N_{opt} \propto C^{0.63 \pm 0.08}$ and
$D_{opt} \propto C^{0.44 \pm 0.06}$. 

As discussed in the introduction, in Figure~\ref{fig:loss_bands} (Left), we observe that the optimal models required for the motion forecasting task are 50 times smaller in the number of parameters than a large-language model at the same compute budget, for example, see~\citet{hoffmann2022training}. Other studies of autoregressive model scaling on non-language domains have observed similar trends, notably~\citet{henighan2020scaling}, where they systematically studied scaling on different modalities. There we observe a 1-2 orders of magnitude difference in the optimal model size between different domains. Language seems to require the most parameters and is trained on less data at a fixed compute. Understanding whether this is a property of the data modality or the data mixture and distribution could improve the data efficiency of our models.
 
\subsection{Cross-Entropy Loss Scaling}
\begin{figure}[H]
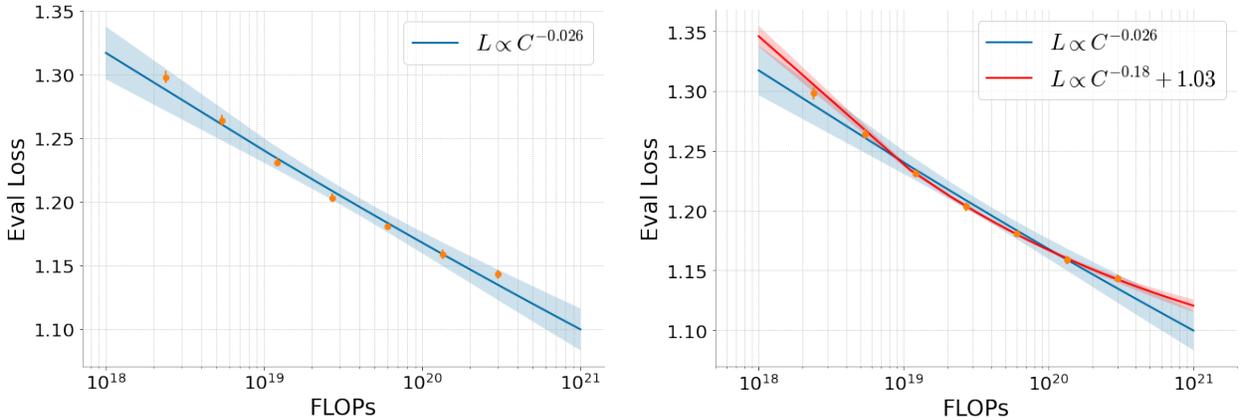

    \centering
    \begin{subfigure}{0.49\textwidth}
        \includegraphics[width=\textwidth]{figures/optimal_loss.pdf}
    \end{subfigure}
    \hfill
    \begin{subfigure}{0.49\textwidth}
        \includegraphics[width=\textwidth]{figures/optimal_loss_with_constant.pdf}
    \end{subfigure}
  \caption{Training compute-optimal loss as a function of total compute budget. The $L_{opt}$ points were extracted from from iso-FLOP bands. (Left) The solid line represents a power-law fit to the points. (Right) We fit the points with a power-law plus a constant, we observe that adding a constant captures the curvature in the data points. The error band indicates 3 standard deviations propagated from the fit parameters covariance (see \Cref{appendix:fits_and_errors}).}
  \label{fig:optimal_loss}
\end{figure}

From the iso-FLOP bands, we obtain the optimal loss $L_{opt}(C)$ for a fixed
compute budget. As illustrated in Figure~\ref{fig:optimal_loss}, the optimal
loss improves as a power-law when we scale the model size, data, and compute in
tandem. This demonstrates that the empirical power-law scaling trend studied in
\citet{kaplan2020scaling} for autoregressive language models, and shown in
\citet{henighan2020scaling} to hold for autoregressive generative models across
a wide range of modalities, also applies to the task of motion forecasting in
the autonomous vehicles domain. However, we also observe a curvature in the data
points, indicating, that we could be close to an irreducible term.

To address the curvature, we add a constant to the power-law fit form:
\begin{equation*}
    L(C) = a C^b + L_{\infty}
\end{equation*}

where $a$, $b$ and $L_{\infty}$ are the fit constants. The right panel of Figure~\ref{fig:loss}
compares the fit with a constant to a pure power-law fit. We observe that adding a constant explains the data significantly better. This raises the question of whether we are close to the epistemic irreducible loss of the problem, i.e., the entropy of the true data distribution, or if it is an artifact of other factors. These factors could include the limited dataset size of this study; training with multiple data passes (epochs)~\citep{muennighoff2023scalingdataconstrainedlanguagemodels}, and overlap between examples~\citep{hernandez2022scalinglawsinterpretabilitylearning}. The geographic data and complexity of driving scenarios mixture used is another area that is interesting to investigate. We also suspect that any lossiness of the interface with the perception system (we only include a limited set of perception features following the Waymo Open Motion Dataset~\citep{Ettinger_2021_ICCV} configuration in this study) could also contribute to this irreducibility. We leave definitive answers to these questions to future studies.

\begin{figure}[h!]
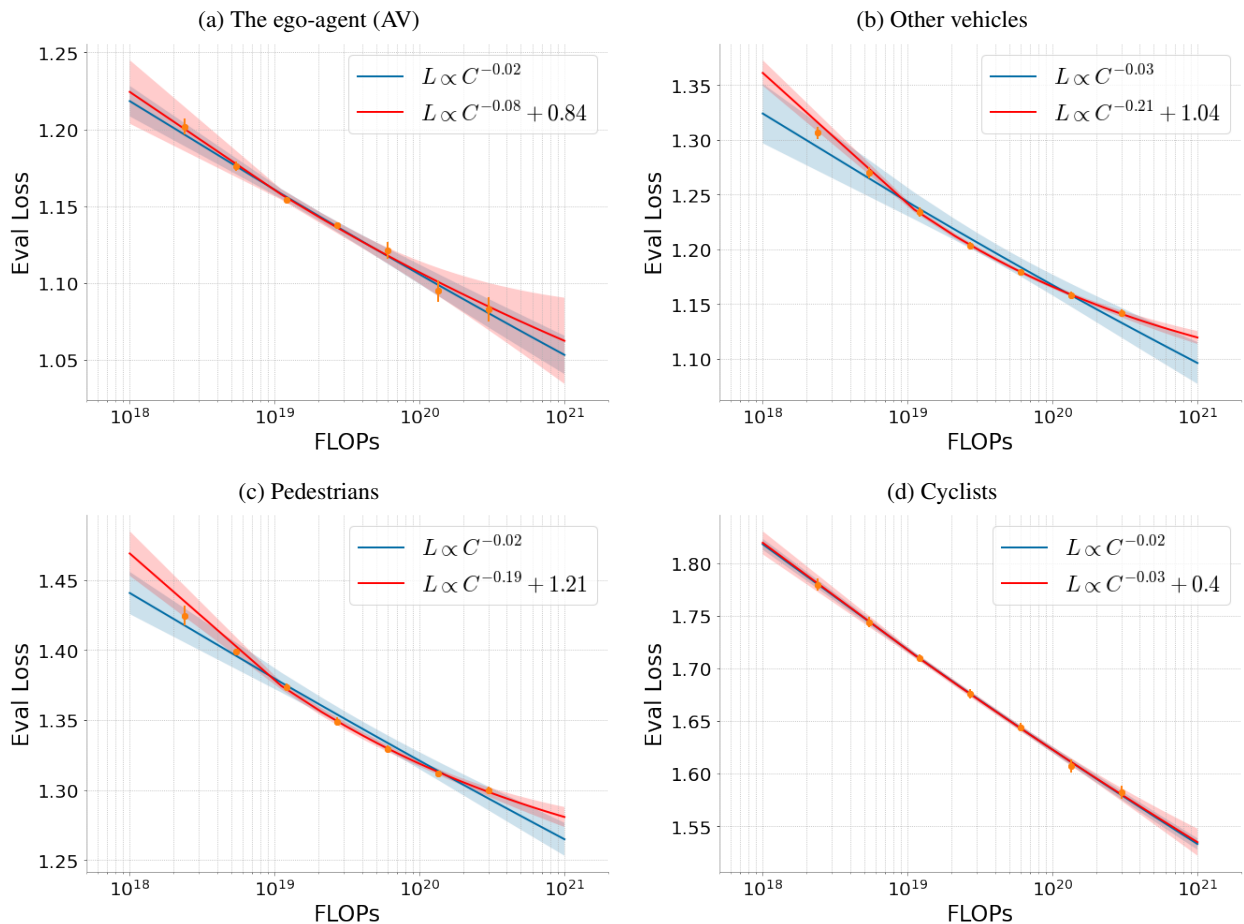

    \centering
    \begin{subfigure}{0.49\textwidth}
        \caption{The ego-agent (AV)}
        \includegraphics[width=\textwidth]{figures/optimal_sdc_loss_with_constant.pdf}
    \end{subfigure}
    \hfill
    \begin{subfigure}{0.49\textwidth}
        \caption{Other vehicles}
        \includegraphics[width=\textwidth]{figures/optimal_veh_loss_with_constant.pdf}
    \end{subfigure}
    \begin{subfigure}{0.49\textwidth}
        \caption{Pedestrians}
        \includegraphics[width=\textwidth]{figures/optimal_ped_loss_with_constant.pdf}
    \end{subfigure}
    \hfill
    \begin{subfigure}{0.49\textwidth}
        \caption{Cyclists}
        \includegraphics[width=\textwidth]{figures/optimal_cyc_loss_with_constant.pdf}
    \end{subfigure}
    \caption{Joint predictions of multiple agents in the scene improves performance for all agents. The slope (exponent) indicates the difficulty of the task. Shown are the the AV, vehicles, pedestrians, and cyclists Eval losses as we scale compute.}\label{fig:optimal_loss_agents}
\end{figure}

Our models are trained on all dynamic agents in the scenes. Different agent types have varying degrees of freedom, speed profiles, and on-road behavior, resulting in different data distributions and predictability. Therefore, studying the trends of these agent types is useful. The main agent types are: the ego-agent (manually driven AV), other vehicles, pedestrians, and cyclists. Top panel of Figure~\ref{fig:optimal_loss_agents} compares the autonomous vehicle (AV) loss to that of other vehicles. Both losses follow a power-law fit, but with different exponents. The scales align with our intuition that AV trajectories are easier to predict due to the lack of perception noise and full observability. The exponents also indicate that improving AV trajectory predictions is easier than for other vehicles, consistent with the latter being a harder task. Figure~\ref{fig:optimal_loss_agents} also shows the losses and exponents for pedestrian and cyclist predictions.

\section{Evaluation}
\label{sec:evaluation}
\subsection{Scaling of Open-loop Distance Metrics}\label{sec:open_loop_metrics}

\begin{figure}
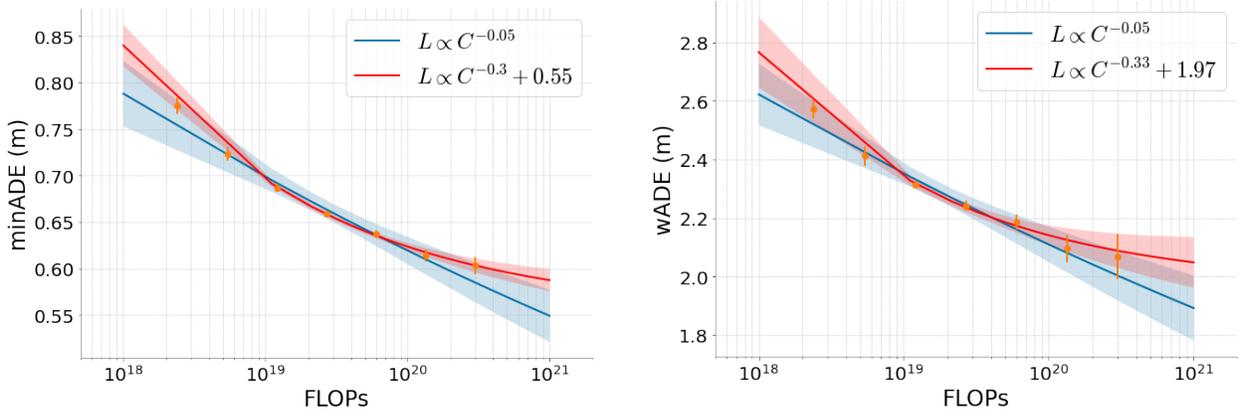

    \centering
    \begin{subfigure}{0.48\textwidth}
        \includegraphics[width=\textwidth]{figures/optimal_minADE_with_constant.pdf}
    \end{subfigure}
    \hfill
    \begin{subfigure}{0.48\textwidth}
        \includegraphics[width=\textwidth]{figures/optimal_wADE_with_constant.pdf}
    \end{subfigure}
    \caption{Open-loop distance metrics, minADE and wADE, of compute-optimal models as a function of the total training compute budget, along with power-law fits, with and without a constant. Open-loop model performance improves as the total training compute budget increases. We show power-law fits to help in discussions, but we caution that establishing a power-law relationship would require a study over many more orders of magnitude of compute.}
    \label{fig:openloop_metrics}
\end{figure}

In this section, we assess how well improvements in cross-entropy loss, driven by scale, correlate with improvements in the model's driving skills using common open-loop distance metrics.

\textbf{Rollout Aggregation} is required to evaluate in standard motion forecasting benchmarks, such as WOMD~\citep{Ettinger_2021_ICCV}. These often require representing predicted trajectories as a small set of distinct "modes" with associated probabilities, per agent. 
These modes can capture different maneuver outcomes (like yielding or passing) or subtle variations in speed and path. To achieve this compact representation, we aggregate 64 predicted trajectories (rollouts) into 12 representative trajectories using NMS and K-means clustering, following ~\citet{multipathpp} and ~\citet{Seff2023MotionLM}.
This technique can be thought of as a soft version of majority voting, a common technique in aggregating samples in LLMs. 
The following metrics then evaluate these 12 clustered trajectories.

\textbf{Minimum Average Displacement Error (minADE)} computes the
    Euclidean distance error of the trajectory closest to the ground truth.
\begin{equation}
    minADE = \min_{k \in 1, \dots, K} \frac{1}{T}\sum_{t=1}^{T}||y_{t}^k - \hat{y}_{t} ||_{2}
\end{equation}

where $y^k$ is the $k$-th predicted trajectory for a given
agent, $\hat{y}$ is the ground truth trajectory for the agent, and T is the
number of time steps per trajectory. The final metric is averaged over
the total number of agents predicted over the evaluation dataset.

\textbf{Weighted Average Displacement Error (wADE)} computes the average distance
error weighted by the probability of each trajectory cluster. This metric is
more sensitive to the model's coverage of different maneuver modes.

\begin{equation}
  wADE =  \frac{1}{T}\sum_{k=1}^{K} p_{k}\sum_{t=1}^{T}||y_{t}^k - \hat{y}_{t} ||_{2}
\end{equation}
where $K$ is the total number of predictions (clusters) per agent, and $p_k$
is the probability of that prediction. Similar to minADE, the final metric
is averaged over the total number of agents predicted over the evaluation
dataset.

Instead of evaluating the optimal models identified in the loss scaling law study, we repeat the iso-FLOP procedure for each metric to obtain their respective scaling trends. 
Figure~\ref{fig:openloop_metrics} displays the power-law fits for both the minADE and wADE metrics.
We observe that the power-law improvement in the loss translates to these open-loop metrics. While we show power-law fits in our figures, we caution that establishing a power-law relationship would require a study spanning many more orders of magnitude of compute and more rigorous statistical methodology~\citep{clauset2009powerlaw}. We use power-law fits here as the simplest hypothesis suggesting that a monotonically decreasing cross-entropy would also minimize the distance of the closest predicted trajectory to the ground truth. In fact, a parabolic form fits our observed data better, but would likely introduce high variance if the study were extended to higher compute. We defer establishing whether an empirical "law" governs the scaling of distance metrics to future studies with larger data and compute budgets.

\begin{figure}[H]
    \centering
    \begin{subfigure}{0.55\textwidth}
        \includegraphics[width=\textwidth]{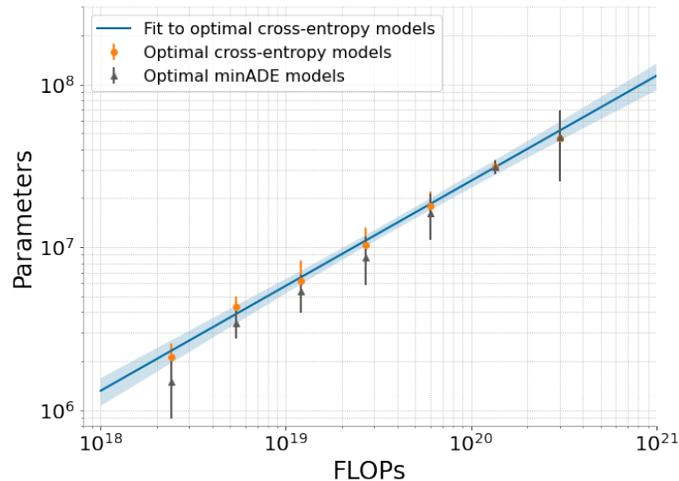}
    \end{subfigure}
    \hfill
    \caption{Comparison of the model number of parameters optimized for Eval cross-entropy versus minADE. For smaller compute budget, minADE-optimal models tend to be smaller than cross-entropy-optimal models and more consistent for larger compute budgets. }\label{fig:optimal_params_loss_and_minADE}
\end{figure}

Figure~\ref{fig:optimal_params_loss_and_minADE} compares the number of parameters of minADE-optimal models to the cross-entropy-optimal models, the two are generally consistent within errors, the minADE-optimal models could be smaller at lower compute budgets. As in the loss study, we use the metric values at the minima of the parabolas to fit power-laws, with and without a constant, as a function of the total compute budget. 

\begin{figure}[H]
    \centering
    \begin{subfigure}{\textwidth}
        \includegraphics[width=\textwidth]{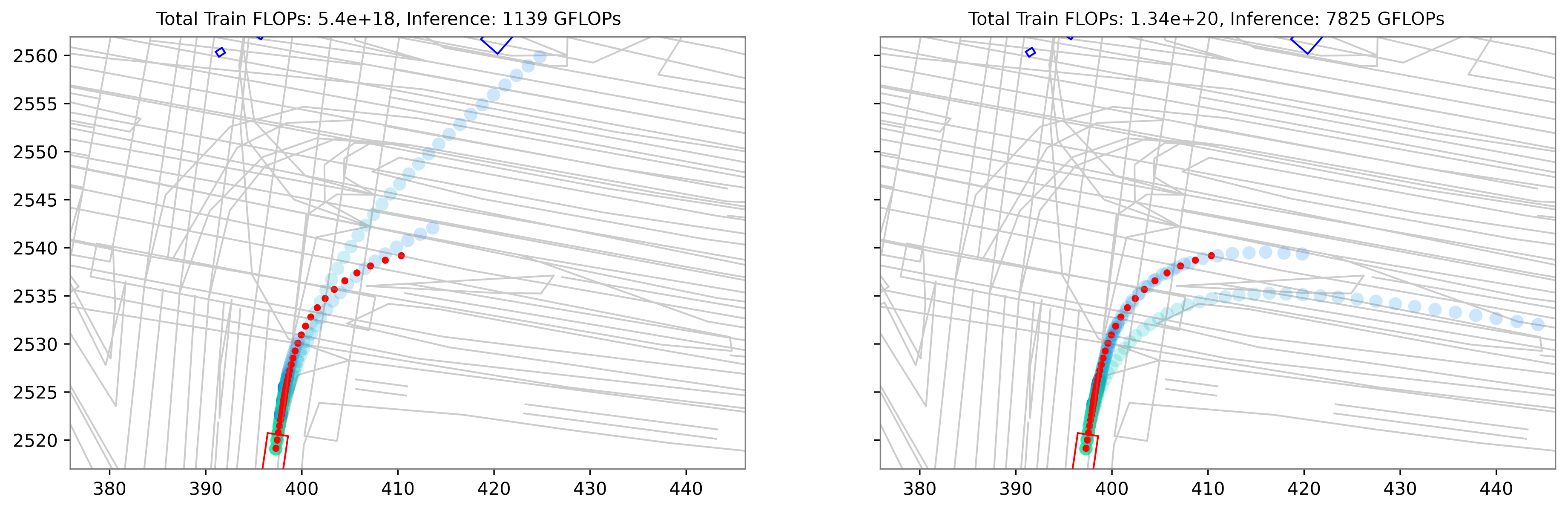}
    \end{subfigure}
    \hfill
    \begin{subfigure}{\textwidth}
        \includegraphics[width=\textwidth]{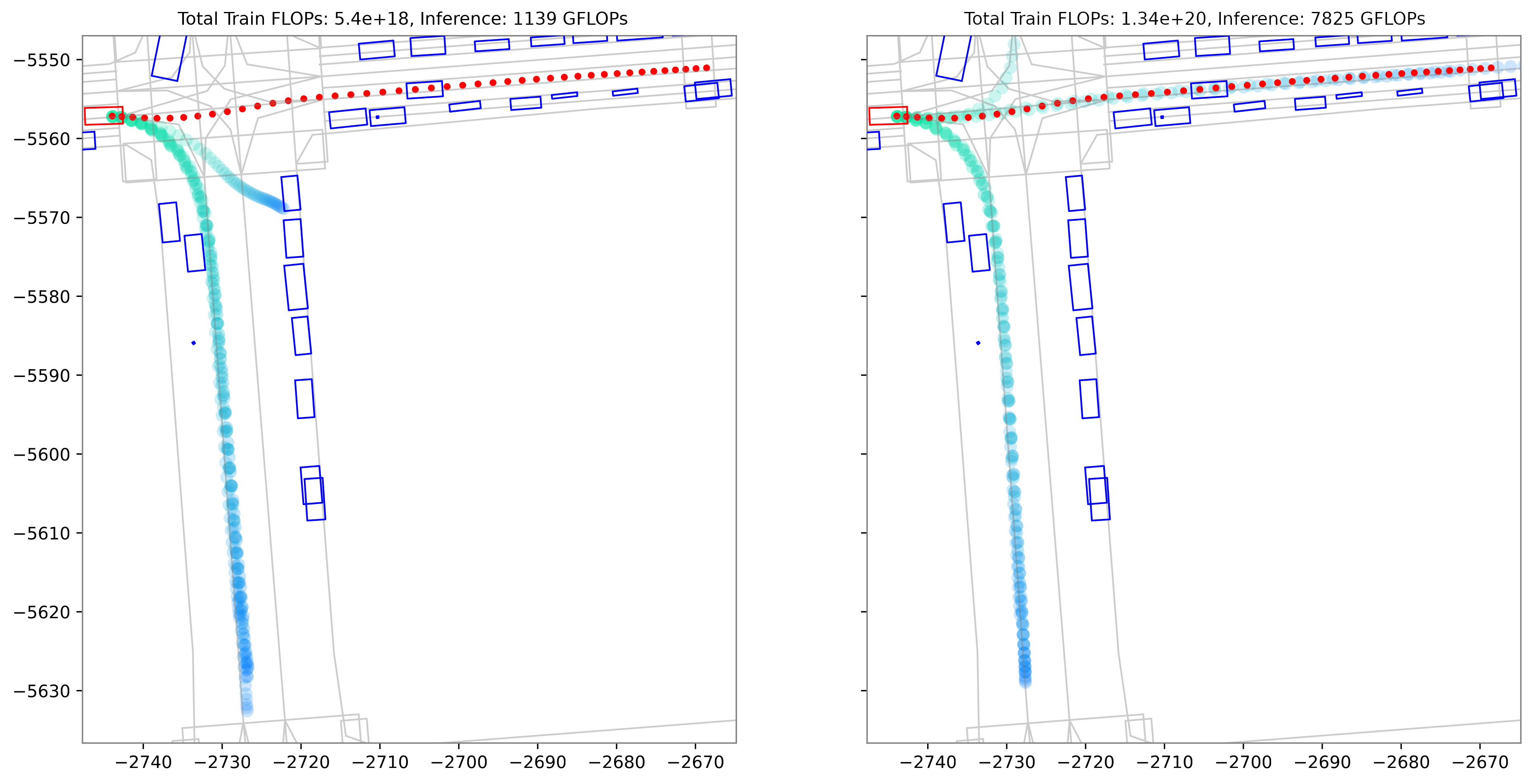}
    \end{subfigure}
    \caption{Qualitative differences between models of increasing size. Blue/green - predicted trajectories, Red - Ground truth trajectory.}\label{fig:qualitative_model_size}
\end{figure}

Figure~\ref{fig:qualitative_model_size} shows the qualitative difference between models of increasing size. These examples show that diversity and accuracy improve as the model size increases. See the Appendix~\ref{appendix:viz_training_scaling} for more visualizations.

\subsection{Closed-loop scaling}

While other domains also show improvements in open-loop model performance from scaling up models,
closed-loop performance is a relatively unexplored area of research. In many domains---and of particular interest, in robotics---it is an active debate whether improvements on open-loop metrics (e.g., cross-entropy or distance from expert demonstration) translate to improvements in closed-loop performance.  Showing that open-loop scaling studies translate to closed-loop performance can be highly promising for safety-critical applications like autonomous vehicles and embodied AI robotics in general.

\begin{figure}[H]
    \centering
    \begin{subfigure}{0.6\textwidth}
        \includegraphics[width=\textwidth]{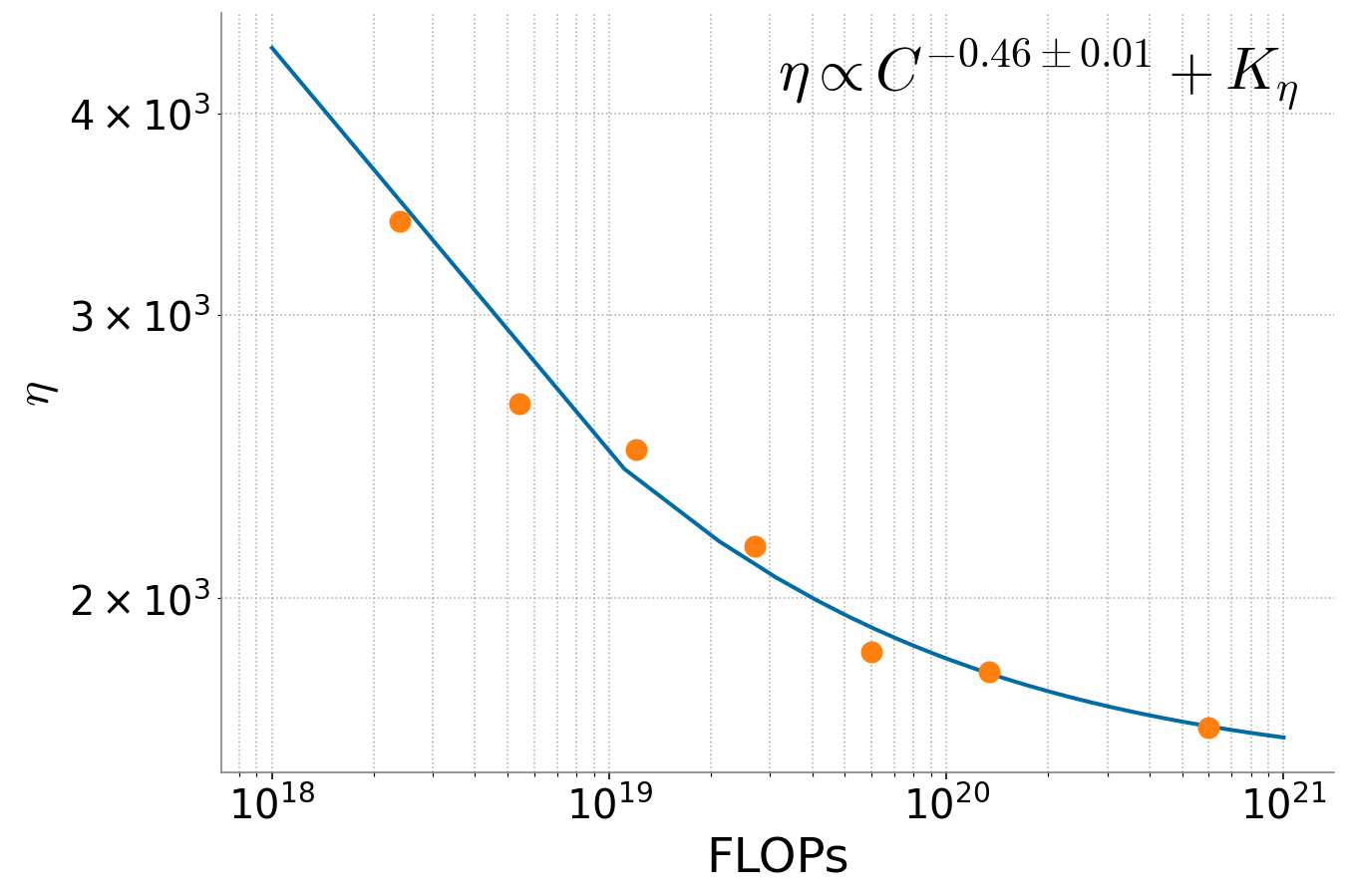}
    \end{subfigure}
    \hfill
    \caption{Performance in closed-loop, as indicated by the number of failures $\eta$, improves as a function of the total pretraining compute budget $C$. We do see similar fits emerge as we see for loss and open loop metrics. $K_{\eta}$ is an irreducible constant.}
    \label{fig:closed_metrics}
\end{figure}

We study closed-loop scaling trends by transferring the scaling series of multi-agent motion forecasting models to an autonomous vehicle (AV) policy that is route-conditioned.
To achieve this transfer, we fine-tune the models to output plans only for the autonomous driving agent ($M$=1), with an additional input of a \textit{planning route}. This planning route is a coarse definition of the path the autonomous vehicle is expected to follow to reach its destination, which we represent as polylines similar to the roadgraph input explained in Section~\ref{sec:model}. All models were fine-tuned to do planning route conditioning using the same compute budget of $10^{15}$ FLOPs. Note that this finetuning is very short compared to pretraining, which is many orders of magnitude larger. Even for the smallest model in the series, this fine-tuning requires 3 orders of magnitude less compute than its pretraining.

In our setup, the closed-loop simulation environment contains a mix of logged agent playback and imitation-learned simulated agents which can interact with the AV planning model. A simulated scenario is run for a fixed time duration of $30$s at a time discretization of $10$Hz.
At every simulated step, we need to execute a single action from the AV policy. To obtain this action, we first compute $R$=128 trajectory rollouts, then select one of the rollouts as a plan $y^*$ according to the equation below, and finally provide only the first 0.1 seconds of the plan as the action to the simulator.
\begin{equation}
  y^* = \argmin_{y_i} \frac{1}{R} \sum_{j} \text{ADE}(y_i, y_j) - \alpha (\text{Progress}(y_i) - \frac{1}{R} \sum_{j} \text{Progress}(y_j))
\end{equation}
where $i \in 1, \dots, R$ and $j \in 1, \dots, R$, $\text{Progress}(y)$ denotes the \textit{progress along the route} of a trajectory, and $\alpha$ is a hyperparameter calibrating the progress bias (higher $\alpha$ indicates a stronger preference for trajectories that progress more).

Calibrating the progress bias is important because different driving policies may have different driving styles in terms of assertiveness. Using a separate closed-loop validation set mined for interestingness and diversity (non-overlapping with the validation set used in previous sections), we tune $\alpha$ for each model in the scaling law analysis such that all models have roughly the same assertiveness. This is done by ensuring that the ratio between the number of scenarios where the policy progresses significantly \textit{more} than manual driving and the number of scenarios where the policy progresses significantly \textit{less} than manual driving is similar across models.

We define our closed-loop metric $\eta$ as the number of failed scenarios, and consider a scenario to fail if it does not imitate well the manual driving according to any of these three conditions: (1) the policy progresses significantly \textit{more} than manual driving, (2) the policy progresses significantly \textit{less} than manual driving, or (3) the policy causes a collision.
Note that even though all policies have been calibrated to have a similar assertiveness, better models will be able to achieve a lower $\eta$ by being closer to the manual driving (i.e., less events of the three types above).

In \Cref{fig:closed_metrics} we show closed-loop results when evaluating the calibrated policies of different sizes in the closed-loop validation set.
We see that similar to the scaling law fits for both loss and open loop metrics, closed loop performance also follows a similar scaling trend, with the number of failures $\eta$ decreasing as a power law when scaling pretraining compute.
This suggests that open loop performance can serve as a good proxy for closed loop.

It is worth comparing this result with previous works that found a lack of transfer between open-loop and closed-loop performance, especially for imitation learning based approaches \cite{casas2021mp3,dauner2023parting,dauner2024navsim,li2024ego,cheng2024rethinking,bouzidi2025closing,li2025hydra}.
One important difference is that those studies compared the performance of different models across open-loop and closed-loop benchmarks. In that setup, there are many confounders that could create differing transfer properties: model architecture, model size, objective function, etc.
Our study, in contrast, is much more controlled. We use a simple architecture with minimal inductive biases, a simple loss, and study the sole effect of scaling up the model compute-optimally.
We are glad to identify scale as a key lever to improve performance in both open-loop and closed-loop. We believe this is an important takeaway for the field of autonomous driving. We hypothesize that having a simple architecture with minimal inductive biases is an important piece to achieve this result.

\section{Inference Scaling Laws}
\label{sec:inference}
Scaling the amount of compute used to train models has dramatically improved
their capabilities, as described in previous sections. Here, we explore
inference compute as another axis for scaling by increasing the number of
generated samples to more accurately represent the trajectory distribution.
Similar have done for LLMs as shown in
\cite{brown2024largelanguagemonkeysscaling,snell2024scaling}. To explore the trade-off between
inference compute and model size, we compute the metrics defined in the Waymo
Open Motion Dataset (WOMD) Motion Prediction Challenge described in \cite{Ettinger_2021_ICCV}
while varying the number of samples generated by the models.  In addition to the
minADE metric defined in \Cref{sec:evaluation}, we use the following three metrics:

\textbf{Minimum Final Displacement Error (minFDE)} is equivalent to the minADE
computed only at step T.

\begin{equation}
    minFDE = \min_{k} ||y_{T}^k - \hat{y}_{T} ||_{2}
\end{equation}

\textbf{Miss Rate} computes the proportion of predictions within tolerance
bounds. A miss is defined as the state when none of the individual $K$
predictions for an agent are within a given lateral and longitudinal threshold
of the ground truth trajectory at a given time $T$.  The thresholds are scaled
with both time and speed following \cite{Ettinger_2021_ICCV}. The miss rate is calculated
as the total number of misses divided by the total number of agents predicted.

\textbf{mAP} computes the mean average precision of predictions bucketed by
behavior.  Following~\cite{Ettinger_2021_ICCV}, ground truth trajectories are grouped
into behavior buckets including straight, straight-left, straight-right, left,
right, left u-turn, right u-turn, and stationary. Using the same definition of
a miss defined above, all misses are assigned a false positive and non-misses a
true positive and stored per bucket. An average precision value is computed
from the P/R curve for each bucket. The final metric is the mean average
precision across all buckets.

The number of trajectories sampled from each model was varied from 8 to 1024
increasing by a multiple of 2.  All metrics were computed using $K=6$ by
applying Non-Maximal Suppression(NMS) clustering on the output trajectories
using the aggregation algorithm, the same as described in \Cref{sec:open_loop_metrics}.

\begin{figure}[H]
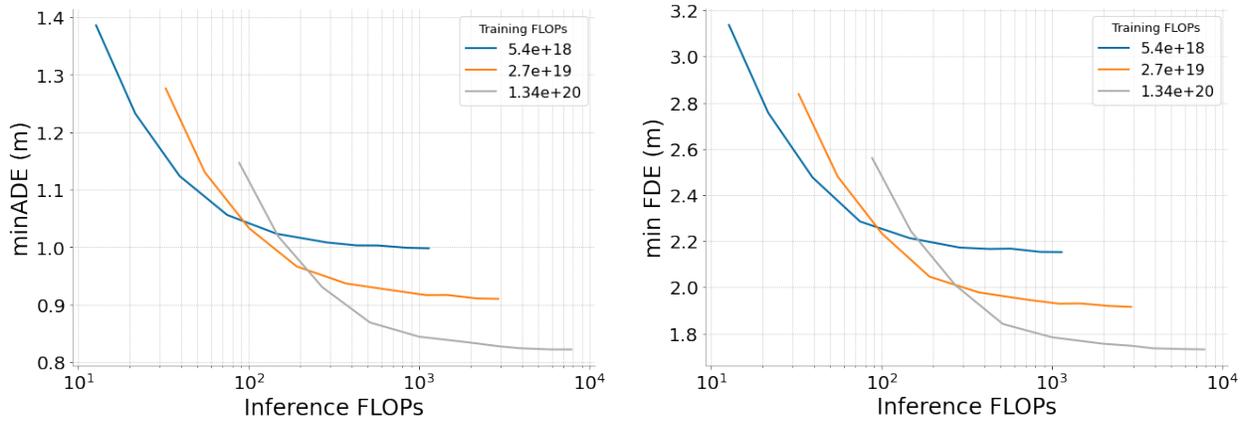

    \centering
    \begin{subfigure}{0.49\textwidth}
        \includegraphics[width=\textwidth]{figures/inference_scaling_min_ade_nms_6_flops.pdf}
    \end{subfigure}
    \hfill
    \begin{subfigure}{0.49\textwidth}
        \includegraphics[width=\textwidth]{figures/inference_scaling_min_fde_nms_6_flops.pdf}
    \end{subfigure}
    \caption{Distance metrics for 3 models of increasing capacity. Each model
    has a range of inference compute for which it is optimal.  There is a
  crossover point for each model at which increasing inference FLOPs results in
a larger model becoming more optimal.}\label{fig:inference_distance}
\end{figure}

\begin{figure}[H]
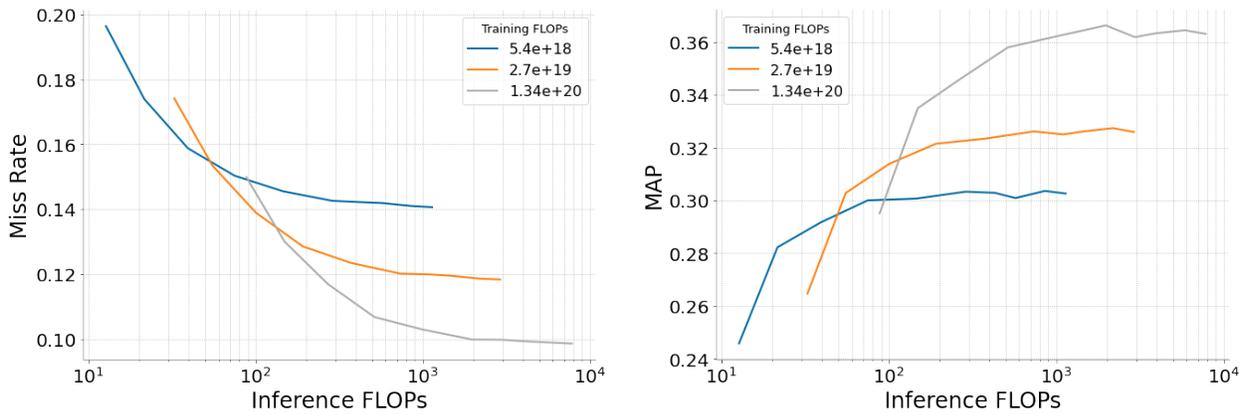

    \centering
    \begin{subfigure}{0.49\textwidth}
        \includegraphics[width=\textwidth]{figures/inference_scaling_miss_rate_nms_6_flops.pdf}
    \end{subfigure}
    \hfill
    \begin{subfigure}{0.49\textwidth}
        \includegraphics[width=\textwidth]{figures/inference_scaling_map_nms_6_flops.pdf}
    \end{subfigure}
    \caption{Coverage metrics for 3 models of increasing capacity. Similar to
    the distance metrics, each model has a range of inference compute for which
  it is optimal.}\label{fig:inference_coverage}
\end{figure}

To simplify the analysis and the plot, we choose three models of increasing capacity to do this analysis. \Cref{fig:inference_distance,fig:inference_coverage} show the computed metrics on these models. We use the pretrained models without any further post-training. We observe that coverage – as measured by the mAP metric – improves as inference compute increases over three orders of magnitude. In addition, distance based metrics like
minFDE/minADE which measure precision, also continue to scale as inference
compute increases. The improvement in each individual model is bounded however,
and increasing inference compute beyond a cross-over point has diminishing
returns. At this point a model with larger capacity becomes more inference
compute optimal. As such, each model has a distinct range over the inference
FLOPs domain for which it is the optimal model as shown in \Cref{fig:inference_distance,fig:inference_coverage}.

\begin{figure}[H]
    \centering
    \begin{subfigure}{\textwidth}
        \includegraphics[width=\textwidth]{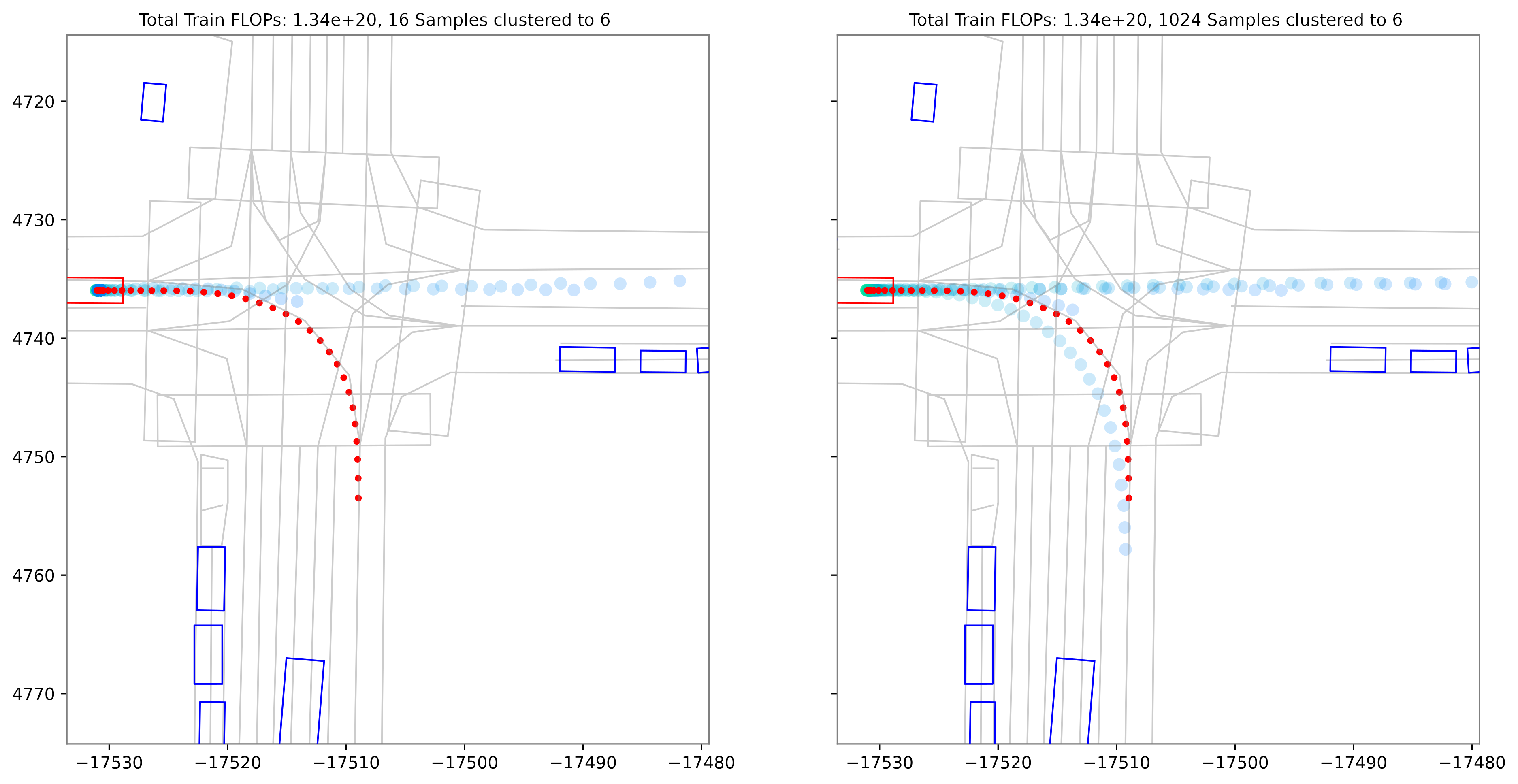}
    \end{subfigure}
    \hfill
    \begin{subfigure}{\textwidth}
        \includegraphics[width=\textwidth]{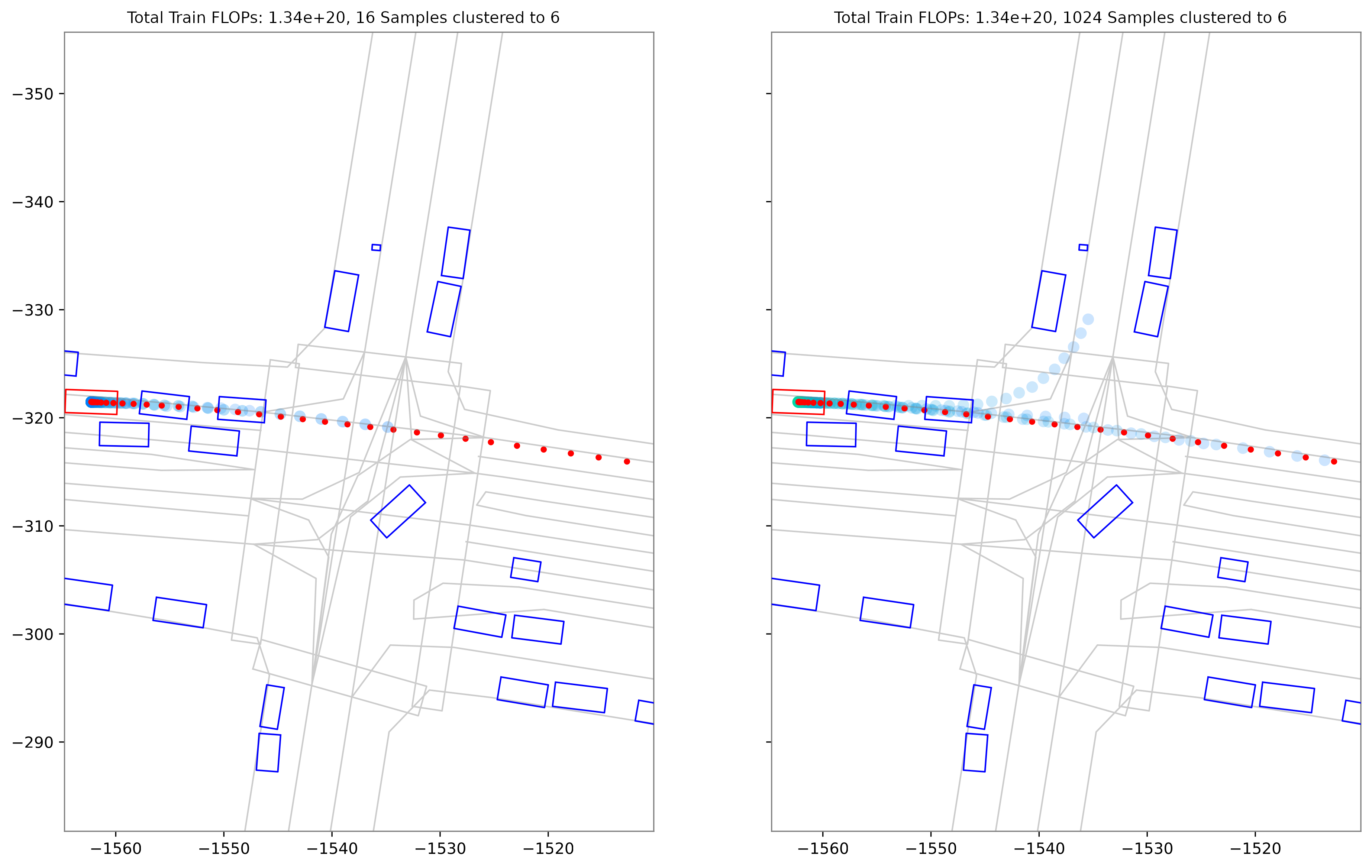}
    \end{subfigure}
    \caption{Qualitative differences between clustered trajectories from 16 and 1024 rollouts. 
    Blue/green - predicted trajectories, Red - Ground truth trajectory.}\label{fig:qualitative_rollouts}
\end{figure}

\Cref{fig:qualitative_rollouts} shows the qualitative difference between clustered predictions from sets of 16 and 1024 samples from the same model. Each set was reduced to 6 trajectories using the method described in the experiments. These examples show that diversity and accuracy improve as the number of samples increases. See the Appendix~\ref{appendix:viz_inference_scaling} for more visualizations.

\section{Cross-agent Skills Transfer}
\label{sec:transfer}
One of the main challenges in advancing machine learning for robotics is the amount of data needed to
train modern, high-capacity deep learning models. Data collection is a slow, human-driven process that
requires platforms, logistics, and time, all of which pose significant challenges to scalability. To
address this challenge, there is a growing body of research focusing on pretraining on passive robotic
video data without robotic-action labels~\citet{ye2024latentactionpretrainingvideos}, and even
pretraining on internet-collected video datasets~\citet{wu2023unleashinglargescalevideogenerative,
cheang2024gr2generativevideolanguageactionmodel}. Similarly, for autonomous vehicle (AV) applications,
determining whether we can leverage videos of observed agents to train AV planner models could help us
train larger foundation models and improve generalizability to new geographic areas without extensive
data collection operations. In this brief study, we aim to shed light on two questions: 1) Does training
on observed trajectories of other agents transfer to the AV agent? 2) If so, how much is a
human-demonstrated mile worth compared to an observed mile?

\begin{figure}[H]
  \centering
  \includegraphics[width=0.7\textwidth]{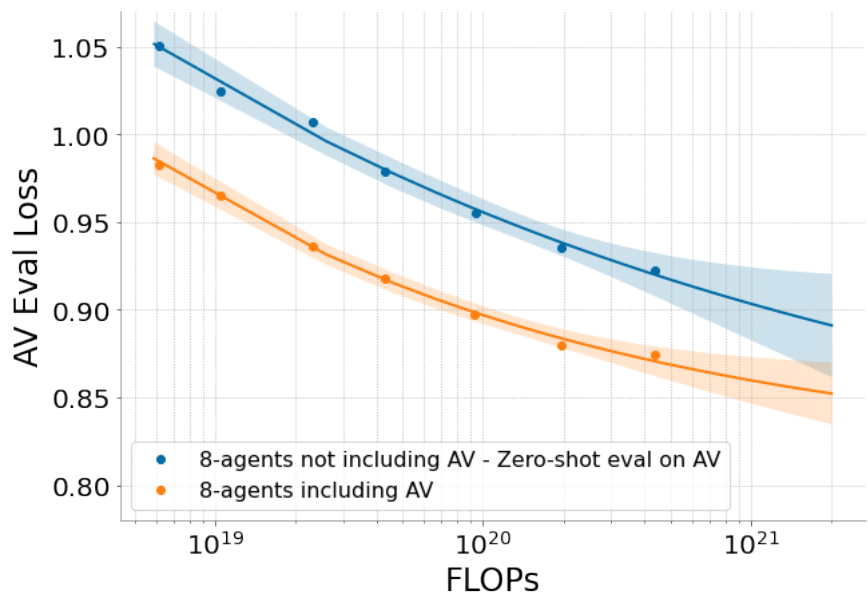}
  \caption{Compute-optimal models trained on the 8-agent prediction task, excluding the AV, and zero-shot evaluated on 8 agents including the AV, compared to models trained with the AV, as a function of training compute. Training on other agents' trajectories zero-shot transfers to the AV distribution.}\label{fig:cross_agent_flops}
\end{figure}

To answer these questions, we trained compute-optimal models from our main study on joint 8-agent prediction, excluding the AV agent. We then zero-shot evaluated these models on 8 agents, including the AV agent. We compared the AV cross-entropy loss from these models to models trained with AV trajectories\footnote{Note that these models were trained at a later time than the main scaling laws run and used a different action space configuration. Thus, the difference in the absolute value of the loss compared to Figure~\ref{fig:optimal_loss_agents}.}. Figure~\ref{fig:cross_agent_flops} compares these losses as a function of training compute FLOPs. It is clear that models trained without the AV exhibit reasonable zero-shot generalization to the AV agent and follow a similar scaling trend as we increase training compute.

\begin{figure}
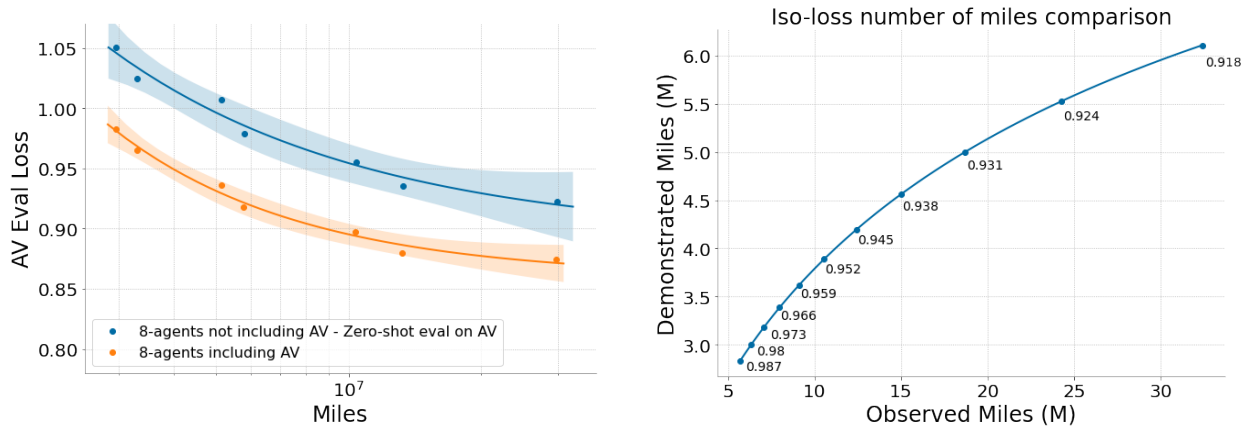

    \centering
    \begin{subfigure}{0.48\textwidth}
        \includegraphics[width=\textwidth]{figures/cross_agent_miles.pdf}
    \end{subfigure}
    \hfill
    \begin{subfigure}{0.48\textwidth}
        \includegraphics[width=\textwidth]{figures/cross_agent_isoloss_miles.pdf}
    \end{subfigure}
    \caption{Compute-optimal models trained on the 8-agent prediction task, excluding the AV, and zero-shot evaluated on 8 agents including the AV, compared to models trained with the AV. (Left) AV cross-entropy loss as a function of training data traveled miles. (Right) Iso-loss comparison: the number of observed miles (excluding AV trajectories in training) needed to achieve the same AV-agent loss as demonstrated miles (including AV trajectories in training).}\label{fig:cross_agent_miles}
\end{figure}

To answer the data equivalency question, we first fit the loss results as a function of the number of training data miles, as shown in Figure~\ref{fig:cross_agent_miles} (Left). We then used these fits to perform an iso-loss analysis, determining the number of observed miles needed to achieve the same loss as demonstrated miles. The comparison is limited to the common loss range between the two curves to reduce extrapolation errors. The trend indicates that every 10 observed miles are equivalent to 2 to 3 demonstrated miles.

One important caveat to this analysis is that all training data was obtained on the same AV-agent platform, so the examples generally correlate with the AV ego-agent. An interesting future direction would be to conduct this study on purely passive observed driving miles or data pooled from different driving platforms. Despite this correlation caveat, we include this analysis to contribute to the discussion of how to address data challenges for robotics applications in general, and as a possible direction to improve the generalizability of AV foundation models to new geographic areas and driving behavior.

\section{Related Work}
\label{sec:related_work}
\textbf{Neural Scaling Laws.} Neural scaling laws have become a critical area of research in deep learning, addressing mission-critical questions such as: Given a finite compute budget, what are the optimal model and dataset size scaling strategies? And how does investment in model scaling translate to tangible model and product improvements? The early work by \citep{hestness2017deep} demonstrated that model accuracy improves as a power law with increased training dataset size across various model architectures in four domains: machine translation, language modeling, image processing, and speech recognition. \citet{kaplan2020scaling}'s systematic study of autoregressive transformer decoder scaling for language modeling revealed that larger models with more parameters, trained on more data and compute, consistently and predictably perform better. They also developed a methodology for determining the optimal model and dataset size as a function of training compute. \citet{henighan2020scaling} investigated cross-entropy loss scaling in four generative modeling domains: image modeling, video modeling, multimodal image-to-text models, and mathematical problem solving. In each case, they showed that joint scaling of model, data, and compute follows a power-law plus constant scaling law. Utilizing the information-theoretic interpretation of cross-entropy loss as a KL-divergence plus the self-entropy of the training data, they suggested that empirical scaling laws could predict both terms. \citet{hernandez2021finetuningscaling} extended these findings to transfer learning, highlighting the importance of scaling pretraining for generalization across tasks and demonstrating that transfer also scales predictably in terms of parameters, data, and compute. Encoder-decoder scaling laws have been studied in \citep{ghorbani2021scaling}. Furthermore, \citet{hoffmann2022training}, known as the Chinchilla study, introduced a methodology for optimizing the optimizer scheduler for best performance at a fixed number of training steps. They subsequently investigated training-compute optimal model scaling using three distinct approaches. One of their significant contributions is the development of the IsoFLOP analysis approach, which we adopt in this study.

\textbf{Scaling laws in AV.} Several studies have investigated scaling laws for autonomous vehicle (AV) models, with emphasis on behavior forecasting and planning. For instance, \citep{hu2023gaia} explored the scaling of a multimodal Vision-Language-Action model incorporating a diffusion video decoder for end-to-end conditional world modeling. Utilizing 4,700 hours of driving data, they examined scaling across four orders of magnitude of compute, demonstrating improved video generation performance with increased model and compute scales. In contrast, our study focuses on the behavior forecasting task and leverages an approximately two orders of magnitude larger driving dataset. Furthermore, we systematically investigate the optimal model and dataset sizes relative to training compute. \citet{nuro2025ai} separately analyzed the scaling of a transformer-based behavior prediction model and a perception model, using an undisclosed training dataset size and a computational budget below $10^{17}$ FLOPs. While they observed improvements in open-loop metrics with increased scale, they reported overfitting in their largest models and cautioned against the benefits of scaling beyond a certain threshold. \citet{zheng2024preliminary} investigated data scaling laws for an end-to-end model family designed for joint online map prediction, 4D occupancy prediction, static detection, motion prediction, and planning. They assembled a dataset of 4 million demonstrations from 23 diverse scenarios, totaling 30,000 hours, to evaluate model performance and generalization. Through both open- and closed-loop evaluations, they demonstrated a power-law relationship between model performance and training data volume. Additionally, they highlighted the significant performance gains achieved by increasing the representation of long-tail scenarios. \citet{huang2024drivegpt} examined the scaling of a transformer model closely aligned with our model and task. Due to computational constraints, they limited the optimal model and data scaling to $10^{15}$ and $3 \times 10^{17}$ FLOPs. They reported improved performance metrics with increased scale and qualitatively demonstrated robust closed-loop performance for their larger models.

\section{Discussion}
\label{sec:discussion}
In this report, we presented our scaling laws study of an encoder-decoder
transformer model for a joint prediction task in the autonomous driving domain.
Like the case for LLMs, we observe that improvements in the cross-entropy loss
follow a power-law as we scale the training compute, while jointly scaling the model and
dataset sizes. However, unlike the case for LLMs, the optimal models for this task
tend to be relatively smaller in size, while requiring significantly more data to
train. We believe that this finding, if it holds for similar robotic planning tasks,
has important implications for data collection and the sizes of models that should be
trained. Furthermore, the smaller sizes of these models result in lower latency,
which implies that improvements in onboard system performance can be directly driven
by scaling training dataset size and compute.

We know that, while driving data is highly multi-modal, the distribution of the
training data is dominated by less interesting modes, like driving straight. We
also hypothesize that, unlike the case for language, driving intuitively
requires less knowledge building and retrieval and more spatial reasoning, so
the optimal models for this planning task would likely have relatively fewer
parameters in the feed-forward network layers. An interesting research
direction is to understand which of these observations could help explain the
relatively smaller sizes of the optimal models.

Another interesting direction is to investigate what changes would make these
models more data-efficient on this task. For example, would sampling more
challenging driving scenarios require spending more compute per example and
help the model learn more from less data? How would using richer perception
inputs, such as vision tokens, or moving to an end-to-end paradigm affect the
data efficiency of these models? On the model output side, would increasing the
complexity of the world-modeling, e.g., the action space, the number of modeled
agents, or the model output modality, make the models learn more from each
example? These are all interesting questions that we hope to see answered as we
learn more about the nature of these empirical scaling laws, especially for
domains beyond language.

While we show that improvements in the cross-entropy loss lead to improvements in
open and closed-loop metrics, our current computational budget does not span enough orders of
magnitude to establish whether they, too, follow a power-law relationship with
training compute. We hope that future studies will help obtain a predictive relation,
if one exists.

We observe that cross-entropy and all metrics are better explained by a power-law plus a constant. It is important for future studies to understand whether this is measuring the epistemic irreducible loss of the problem, i.e., the entropy of the true data distribution, or if is an artifact of other factors. Our leading hypotheses are that the limited dataset size, systematic duplication of features by overlapping examples, and the basic set of scene-level perception features used are the leading contributors to the observed constants.

The improvement of models by scaling inference-time compute is a particularly
interesting result. First, we see that improving models only by increasing sampling
has a point of diminishing returns, at which point it is better to sample a larger
and more expensive model, confirming the need for training larger models with larger
compute budgets and datasets. Second, the improvement itself calls for more adaptive
strategies for how to choose inference-time compute to solve more challenging
scenarios. There are recent proposals for how to do this for
LLMs~\cite{snell2024scaling}. Similarly, for planning tasks in robotics, different
scenarios vary drastically in their complexity, and onboard systems could adapt the
compute needed to solve them.

There is also the question of how to train inference-time compute-optimal
models~\cite{devries2023chinchilla_analysis}. These are smaller models trained past
the Chinchilla-optimal point, which makes them cheaper for onboard applications.
There are proposals for directly optimizing to find the corresponding optimal model
and training dataset sizes~\cite{sardana2023beyond}. In the future, we can also
investigate adding the number of samples per prompt as another parameter in the
optimization. Another exciting direction would be, instead of
training a suite of models of different sizes, to train a nested architecture that
allows us to elastically choose the model capacity to use at inference
time~\cite{kudugunta2023matformer}. It would be very interesting to see how to
combine such architectures with more adaptive test time sampling techniques.

Finally, an important question for the field of robotic manipulation is devising scalable data collection methodologies that can unlock the pre-training of larger foundation models. We hypothesize that for driving, one such source can be passively observed driving logs. While all of our data is collected from our platform, so it is not truly passive, nevertheless, we conduct a brief study to investigate whether training on observed trajectories of other agents zero-shots transfers to the AV ego-agent. We hope this study can help spur more research in this area.

We believe this is a very rich area of research for robotics, and we are looking
forward to learning more about the model scaling laws for modalities and tasks
beyond natural language.

\section{Acknowledgements}
\label{sec:acknowledgements}
This work represents a significant collaborative effort, and we wish to extend our sincere gratitude to the many colleagues who provided critical support. Their invaluable contributions spanned all stages of the project, including data curation, infrastructure support, experimental validation, and insightful feedback. 

We would like to especially acknowledge Brian Axelrod, Anand Bhaskar, Aaron Effron, Aditya Gupta, Wenlin Hu, Gus Katsiapis, Lingyun Li, Maher Mneimneh, Cheol Park, Rushina Shah, Avikalp Srinivasta, Thomas Tian, and Daniel Wang.

\pagebreak

\bibliographystyle{abbrvnat}
\bibliography{bibliography}

\appendix
\appendixpage
\addappheadtotoc
\label{sec:appendices}
\section{FLOPs and Parameters Computation}
\label{appendix:flops}

When computing flops and number of parameters, we include einsums in attention and feed-forward layers. We ignore embedding and normalization layers. When computing flops, each multiply-add counts as 2 flops by cost.

Flops and number of parameters are functions of the following hyper-parameters:

\begin{itemize}
    \item $\textbf{E}$: number of scene tokens
    \item $\textbf{n}$: number of encoder layers
    \item $\textbf{D}$: number of decoder query tokens
    \item $\textbf{m}$: number of decoder layers
    \item $\textbf{d}$: transformer hidden dimension
\end{itemize}

Throughout our experiments, we use $4d$ as the feed-forward layer hidden dimension.

Flops are computed as:

\begin{itemize}
    \item \textbf{Single encoder layer}: $ 24 E d^2 + 4 d E^2 $
    \begin{itemize}
        \item \textbf{Scene self attention}: $ 8 E d^2 + 4 d E^2 $
        \item \textbf{Feed-forward layer}: $ 16 E d^2 $
    \end{itemize}
    \item \textbf{Single decoder layer}: $28 D d^2 + 4 d D^2 + 4 E d^2 + 4 d D E$
    \begin{itemize}
        \item \textbf{Query self attention}: $ 8 D d^2 + 4 d D^2 $
        \item \textbf{Query-scene cross attention}: $ 4 D d^2 + 4 E d^2 + 4 E D d $
        \item \textbf{Feed-forward layer}: $ 16 D d^2 $
    \end{itemize}
    \item \textbf{Total flops}: $n * (24 E d^2 + 4 d E^2) + m * (28 D d^2 + 4 d D^2 + 4 E d^2 + 4 d D E) $
\end{itemize}

\textbf{Number of parameters}: $ (12 n + 16 m) * d^2 $

\section{Fits Error Propagation}
\label{appendix:fits_and_errors}
The key question in error propagation is that if one has a function $f(x; \theta_0, \ldots, \theta_n)$, that depends on estimated parameters $\hat{\theta_i}$, and one also has access to their estimated covariance $\sigma_{i,j} = cov(\hat{\theta_i}, \hat{\theta_j})$, how does one estimate the errors in $f(x;, \hat{\theta_0}, \ldots, \hat{\theta_n})$? The standard error propagation technique is to use a first order truncated Taylor expansion of the function around the parameters $\hat{\theta_i}$~\citet{james2006statistical}. A caveat that the truncation produces biased estimates for non-linear functions. Instead or preproducing the expansion here, we explicitly report the formulas used obtain the error bands in our work.

For a function $f(x; a, b, c)$ of three fit parameters $a$, $b$, and $c$, the estimated errors in $f(x)$ as a function of the estimated parameters covariance matrix $\sigma_{i,j}$ is:
\begin{equation}\label{eq:err_prop}
\sigma_f^2 = 
  \left( \frac{\partial f}{\partial a} \right)^2 \sigma_a^2 
  + \left( \frac{\partial f}{\partial b} \right)^2 \sigma_b^2
  + \left( \frac{\partial f}{\partial c} \right)^2 \sigma_c^2
  + 2 \frac{\partial f}{\partial a} \frac{\partial f}{\partial b} \sigma_{ab}
  + 2 \frac{\partial f}{\partial a} \frac{\partial f}{\partial c} \sigma_{ac}
  + 2 \frac{\partial f}{\partial b} \frac{\partial f}{\partial c} \sigma_{bc}
\end{equation}

\subsection*{Parabolic fits}
We fit parabolas of the form 
\begin{equation*}
f(x) = a\left(x - b\right)^2 + c
\end{equation*}
where $a$, $b$ and $c$ are fit parameters. We substitute the following derivatives in equation~\ref{eq:err_prop} to obtain $\sigma_f^2$:
\begin{align*}
\frac{\partial f}{\partial a} &= (x - b)^2 \\
\frac{\partial f}{\partial b} &= -2a(x - b) \\
\frac{\partial f}{\partial c} &= 1.
\end{align*}
We report $\pm 3\sigma_f$ error bands in the figures.

\subsection*{Power-law fits}
We fit power-laws of the form
\begin{equation*}
f(x) = ax^b + c
\end{equation*}
where $a$, $b$ and $c$ are fit parameters. We substitute the following derivatives in equation~\ref{eq:err_prop} to obtain $\sigma_f^2$:
\begin{align*}
\frac{\partial f}{\partial a} &= x^b \\
\frac{\partial f}{\partial b} &= ax^b\ln{x} \\
\frac{\partial f}{\partial c} &= 1.
\end{align*}
We report $\pm 3\sigma_f$ error bands in the figures.

\section{Training Scaling Visualization}
\label{appendix:viz_training_scaling}
The following figures compare the predictions from models of different size with a constant number of samples drawn at inference.
Blue/green - predicted trajectories, Red - Ground truth trajectory.
\begin{figure}[H]
    \centering
    \begin{subfigure}{\textwidth}
        \includegraphics[width=\textwidth]{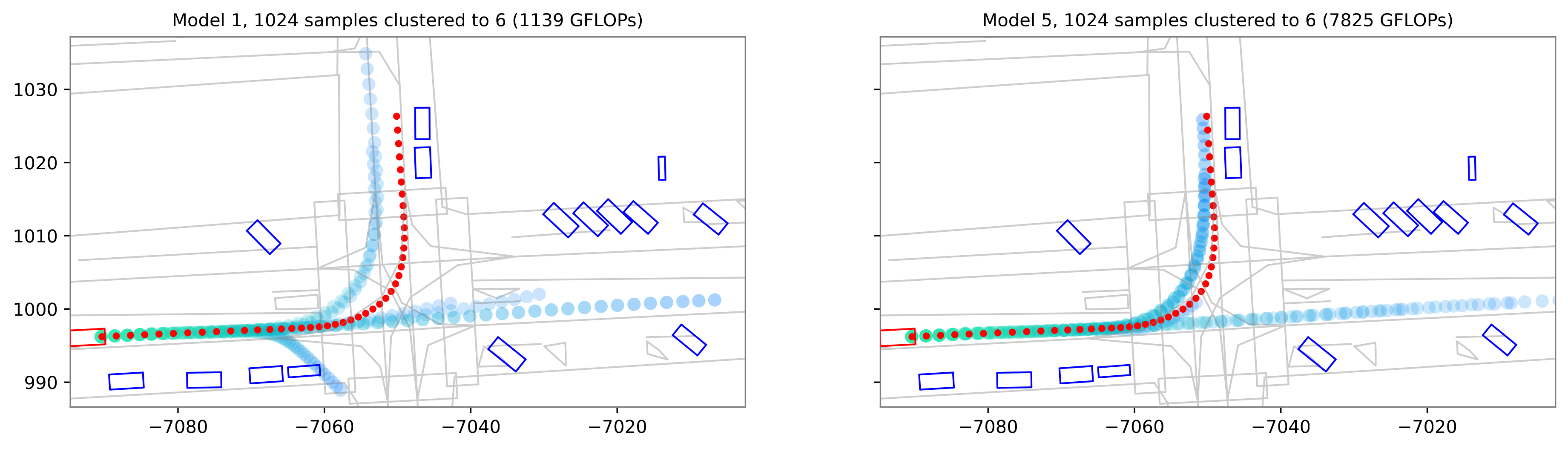}
    \end{subfigure}
    \hfill
\end{figure}
\begin{figure}[H]
    \centering
    \begin{subfigure}{\textwidth}
        \includegraphics[width=\textwidth]{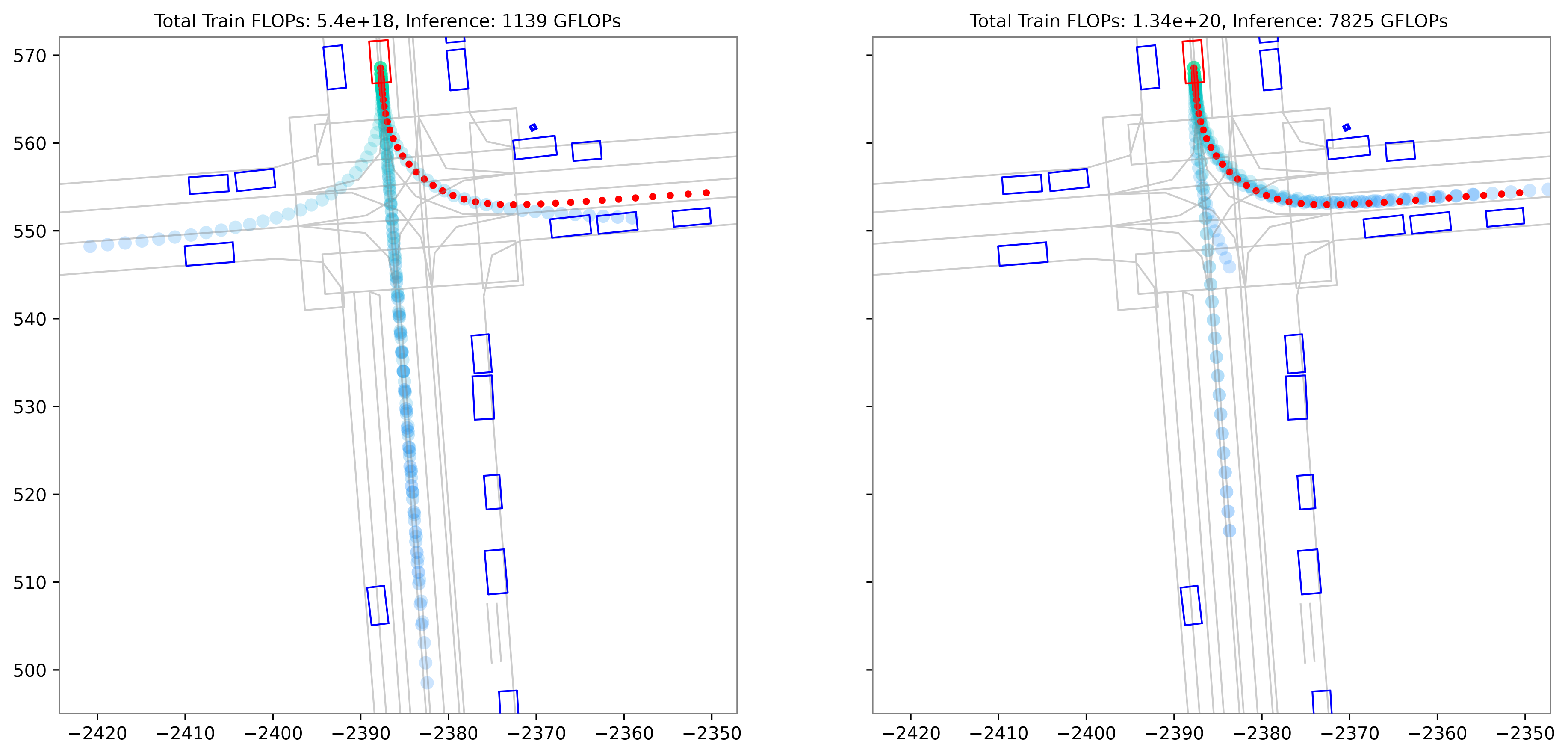}
    \end{subfigure}
    \hfill
\end{figure}
\begin{figure}[H]
    \centering
    \begin{subfigure}{\textwidth}
        \includegraphics[width=\textwidth]{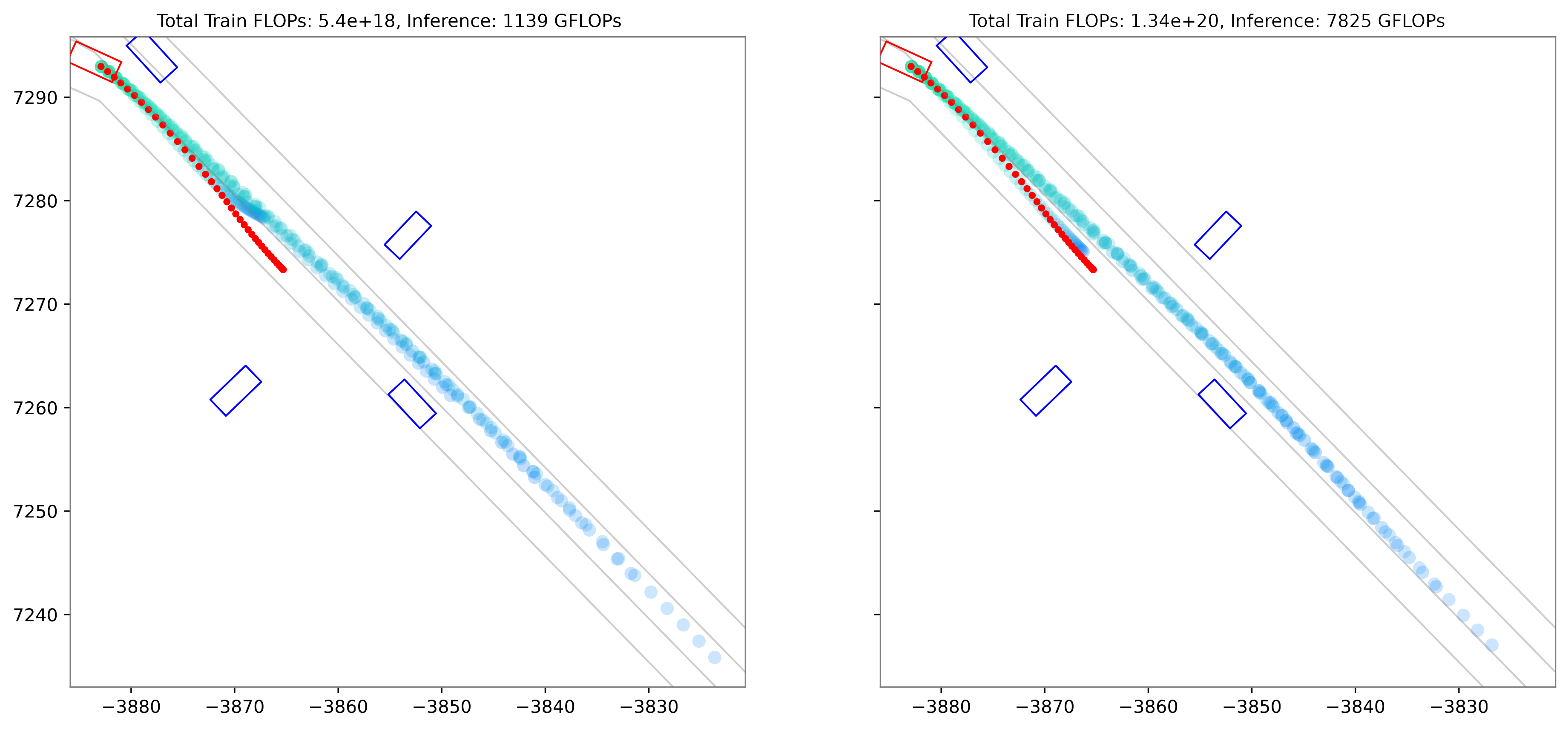}
    \end{subfigure}
    \hfill
\end{figure}    
\begin{figure}[H]    
    \begin{subfigure}{\textwidth}
        \includegraphics[width=\textwidth]{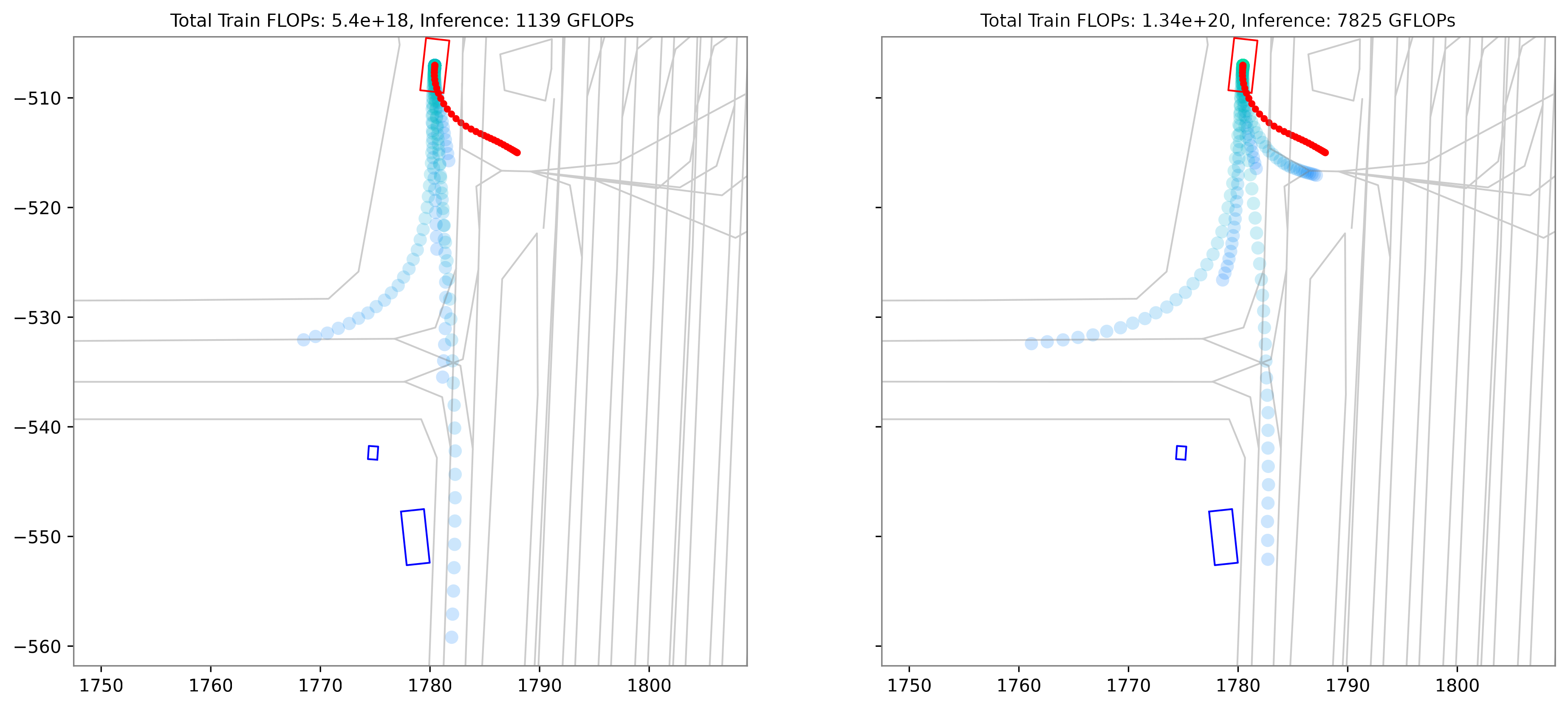}
    \end{subfigure}\label{fig:qualitative_model_size_appendix}
\end{figure}

\section{Inference Scaling Visualization}
\label{appendix:viz_inference_scaling}
The following figures compare the predictions from a single model, varying the number of samples drawn at inference from 16 to 1024. 
Blue/green - predicted trajectories, Red - Ground truth trajectory.

\begin{figure}[H]    
    \centering
    \begin{subfigure}{\textwidth}
        \includegraphics[width=\textwidth]{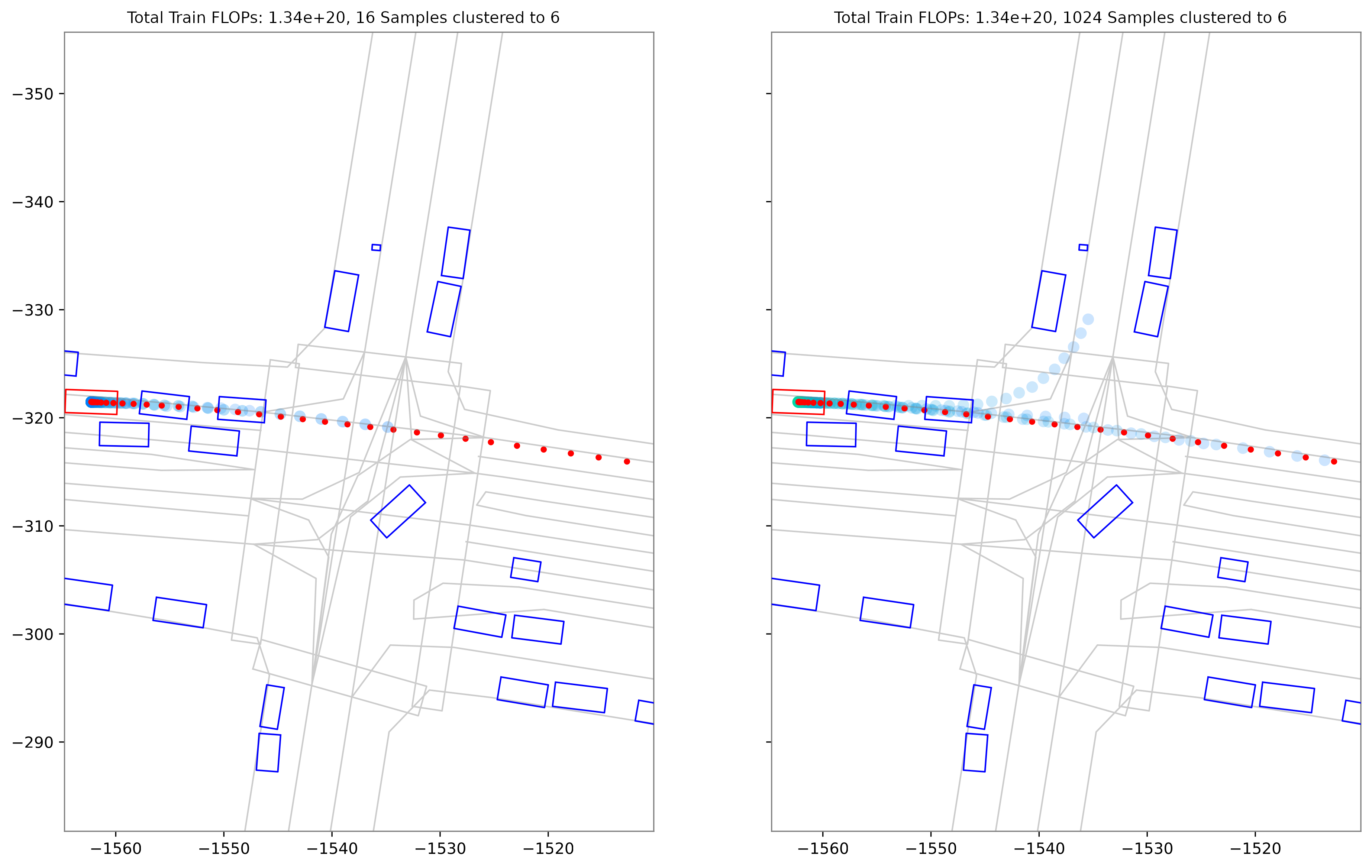}
    \end{subfigure}
\end{figure}
\begin{figure}[H]
    \begin{subfigure}{\textwidth}
        \includegraphics[width=\textwidth]{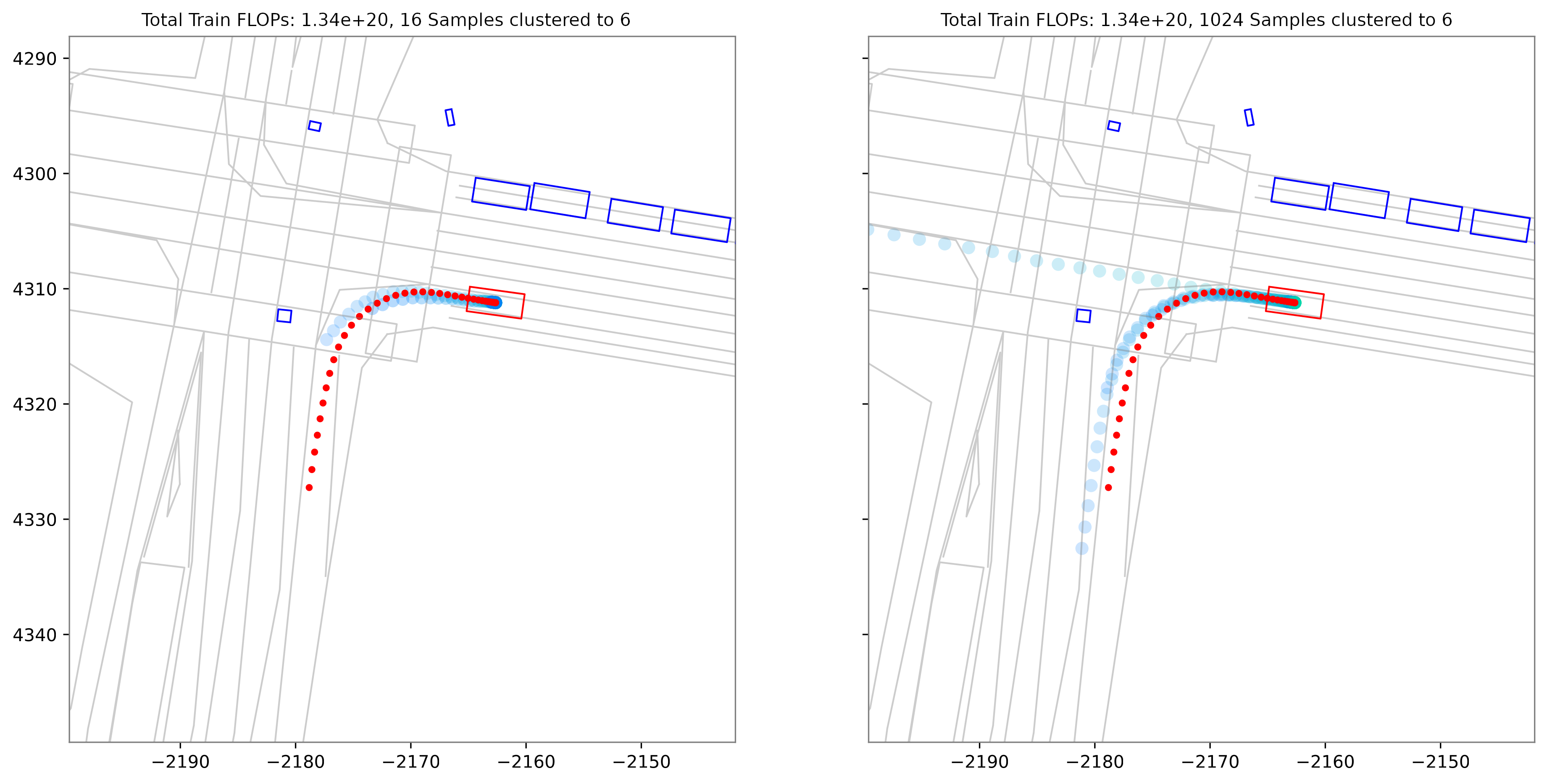}
    \end{subfigure}
    \hfill    
\end{figure}
\begin{figure}[H]
    \centering
    \begin{subfigure}{\textwidth}
        \includegraphics[width=\textwidth]{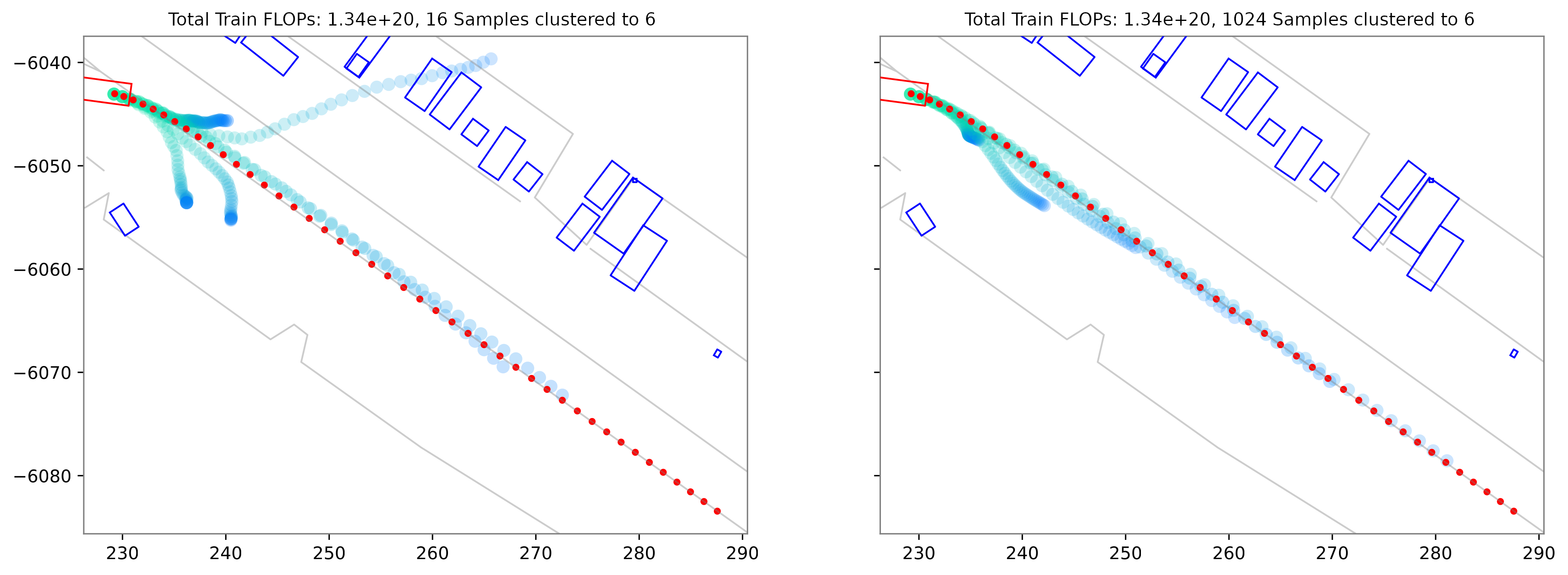}
    \end{subfigure}
\end{figure}
\begin{figure}[H]
    \centering
    \begin{subfigure}{\textwidth}
        \includegraphics[width=\textwidth]{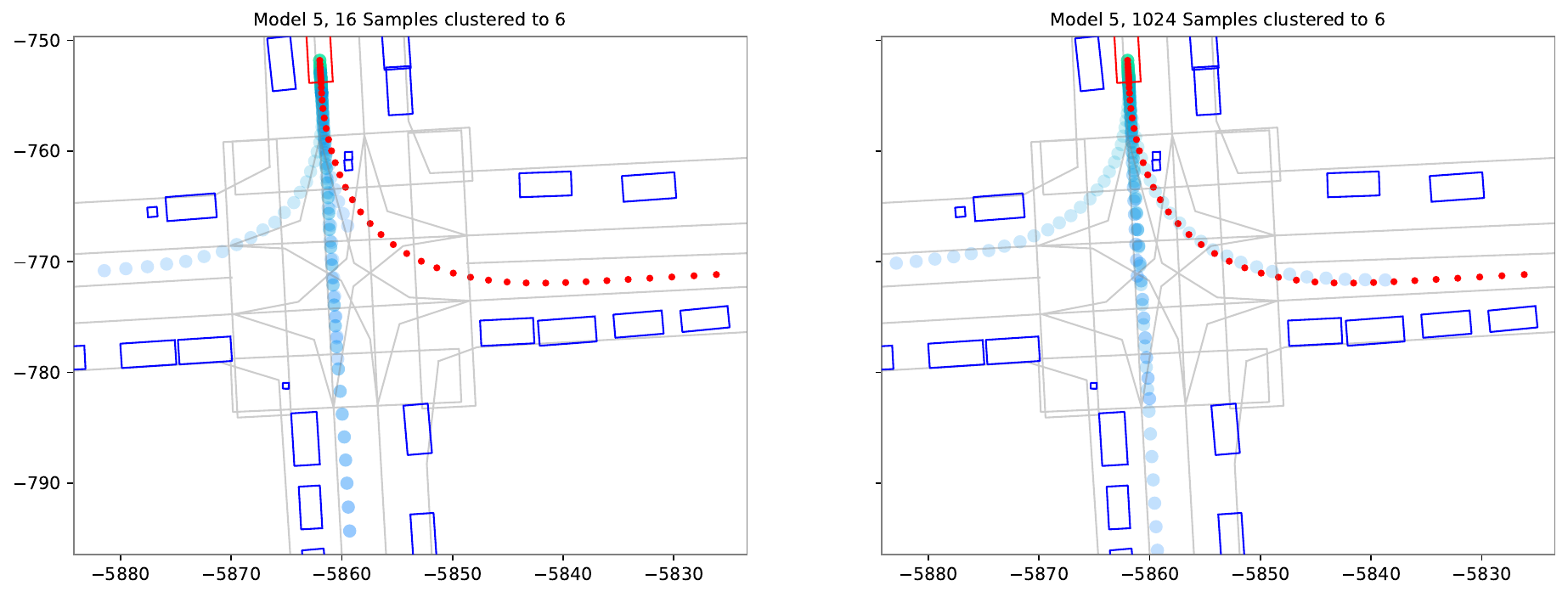}
    \end{subfigure}\label{fig:sappendix_qualitative_rollouts}    
\end{figure}

\end{document}